\documentclass{article}

\usepackage[margin=1in]{geometry}
\usepackage{hyperref}       
\usepackage{url}            
\usepackage{booktabs}       
\usepackage{amsfonts}       
\usepackage{nicefrac}       
\usepackage{microtype}      

\usepackage{graphicx}
\usepackage{subfigure}
\usepackage{amsmath}
\usepackage{amssymb}
\usepackage{amsthm}
\usepackage{bm}
\usepackage{natbib}

\newtheorem{proposition}{Proposition}

\newtheorem{assumption}{Assumption}

\title{Understanding Graph Neural Networks from Graph Signal Denoising Perspectives}

\author{%
  Guoji Fu \and Yifan Hou \and Jian Zhang \\
  \and Kaili Ma \and Barakeel Fanseu Kamhoua \and James Cheng \\
 \date{The Chinese University of Hong Kong}
}

\begin{document}

\maketitle

\begin{abstract}

Graph neural networks (GNNs) have attracted much attention because of their excellent performance on tasks such as node classification. However, there is inadequate understanding on how and why GNNs work, especially for node representation learning. This paper aims to provide a theoretical framework to understand GNNs, specifically, spectral graph convolutional networks and graph attention networks, from graph signal denoising perspectives. Our framework shows that GNNs are implicitly solving graph signal denoising problems: spectral graph convolutions work as denoising node features, while graph attentions work as denoising edge weights. We also show that a linear self-attention mechanism is able to compete with the state-of-the-art graph attention methods. Our theoretical results further lead to two new models, GSDN-F and GSDN-EF, which work effectively for graphs with noisy node features and/or noisy edges. We validate our theoretical findings and also the effectiveness of our new models by experiments on benchmark datasets. The source code is available at \url{https://github.com/fuguoji/GSDN}.

\end{abstract}
\section{Introduction}\label{section1}

Graph data are ubiquitous today and examples include social networks, semantic web graphs, knowledge graphs, and telecom networks. However, graph data are not in the Euclidean space, which renders learning on graph data using traditional machine learning algorithms less effective~\citep{DBLP:journals/corr/abs-1806-01261,DBLP:journals/corr/abs-1901-00596}. To this end, graph neural networks (GNNs) have been proposed as specialized neural network models for learning on graph data. GNNs achieve excellent performance for many important tasks, especially for semi-supervised node classification~\citep{DBLP:conf/iclr/KipfW17,DBLP:conf/nips/HamiltonYL17,DBLP:conf/iclr/VelickovicCCRLB18}.


GNNs are often categorized as \textit{spectral approaches} and \textit{non-spectral approaches}~\citep{DBLP:journals/corr/abs-1901-00596}. Spectral approaches, i.e., \textbf{spectral graph convolution networks} (\textbf{SGCNs})~\citep{DBLP:conf/nips/DefferrardBV16,DBLP:conf/iclr/KipfW17,DBLP:conf/icml/WuSZFYW19}, are mainly inspired by \textbf{graph signal processing} (\textbf{GSP})~\citep{DBLP:journals/tsp/SandryhailaM13}. SGCNs define graph convolution operators as the multiplication of a graph signal with a graph filter in the Fourier domain. Non-spectral approaches, which include \textbf{spatial graph neural networks} (\textbf{SGNNs})~\citep{DBLP:conf/nips/HamiltonYL17,Fey/Lenssen/2019} and \textbf{graph attention networks} (\textbf{GANNs})~\citep{DBLP:conf/iclr/VelickovicCCRLB18,DBLP:journals/corr/abs-1803-03735}, are mainly motivated by convolution operators~\citep{lecun1995convolutional,DBLP:journals/nature/LeCunBH15} or attention mechanisms~\citep{DBLP:journals/corr/BahdanauCB15,DBLP:conf/nips/VaswaniSPUJGKP17} used in computer vision. Similar to convolutions that are defined based on nearby pixels within a window in an image, SGNNs define graph convolutions based on nodes' spatial relations and aggregate the node features of the neighbors of a node within a window. GANNs utilize graph attentions to learn the connection strength of nodes that are connected by an edge, which work similarly to attentions that are used to learn the aggregation weights of pixels.


In spite of GNNs' remarkable performance for representation learning, there has been inadequate work to explain how and why GNNs work so effectively. Recent attempts to explain the working mechanisms of GNNs are either too coarse-grained on the whole graph scale or too restricted for specific GNN models only, e.g., GCN~\citep{DBLP:conf/iclr/KipfW17}. Besides, spectral approaches and non-spectral approaches are studied and explained in different, separate theoretical frameworks, but a unified framework to understand them has been lacking. For example, \cite{DBLP:conf/iclr/XuHLJ19} demonstrate that GNNs are at most as powerful as the Weisfeiler-Lehman test in distinguishing the whole graph structures, \cite{DBLP:conf/aaai/LiHW18} reveal that GCN is conducting Laplacian smoothing on node features, \cite{DBLP:conf/icml/WuSZFYW19} and \cite{DBLP:journals/corr/abs-1905-09550} show that the graph convolution of GCN is a low-pass filter, and \cite{DBLP:conf/nips/YingBYZL19} use two reasoning tasks to empirically show that the strength of graph attentions comes from their ability to generalize to more complex or noisy graphs at test time. However, existing work does not focus on understanding the reasons behind the performance of GNNs (including spectral approaches and non-spectral approaches) for node representation learning.


In this paper, we aim to provide a unified theoretical framework to understand how and why spectral and non-spectral approaches, specifically SGCNs and GANNs, work for node representation learning from \textbf{graph signal denoising} (\textbf{GSD}) perspectives. Our framework reveals that \textit{GNNs are implicitly solving GSD problems}. Specifically, we found that: (1)~\textit{the graph convolutions of SGCNs}, e.g., ChebyNet~\citep{DBLP:conf/nips/DefferrardBV16}, GCN~\citep{DBLP:conf/iclr/KipfW17}, and SGC~\citep{DBLP:conf/icml/WuSZFYW19}, \textit{work as denoising and smoothing node feature}s; and (2)~\textit{the graph attentions of GANNs}, e.g., GAT~\citep{DBLP:conf/iclr/VelickovicCCRLB18} and AGNN~\citep{DBLP:journals/corr/abs-1803-03735}, \textit{work as denoising edge weights}. Based on the theoretical findings, we further design two new GNN models, \textbf{GSDN-F} and \textbf{GSDN-EF}, which conduct effective node representation learning on graphs with noisy node features and/or noisy edges by working through a tradeoff between node feature denoising and smoothing.


Our empirical results on benchmark graphs (with/without noise) validate that the performance of SGCNs on node classification indeed benefits from their ability to denoise and smooth node features. When there is little noise in node features, the node classification performance of SGCNs mainly comes from their ability of node feature smoothing. However, when there is more noise, denoising contributes greater to their performance. The results of node classification on graphs with noisy edges show that the performance of GANNs benefits from their ability to denoise edge weights. Moreover, the results also demonstrate that our proposed models, GSDN-F and GSDN-EF, achieve comparable performance with the state-of-the-art GNNs on graphs with little noise and become more effective on graphs having more noise. The superior performance of GSDN-EF also verifies that its linear edge denoising mechanism can compete with the state-of-the-art graph attentions.
\section{Preliminary and Background}\label{app:background}

We first define the notations used in this paper and introduce the background of graph signal denoising, spectral graph convolutions, and graph attentions.

As in previous works~\citep{DBLP:conf/iclr/KipfW17,DBLP:conf/icml/WuSZFYW19}, we introduce GNNs in the context of node classification. In this paper, the input of a GNN is a graph with node features and some node labels, the output is the predicted labels for unlabeled nodes. Let $\mathcal{G} = (\mathcal{V}, \mathbf{A})$ be a graph. $\mathcal{V}$ is a set of $N$ nodes $\{v_1, \dots, v_N\}$ and $\mathbf{A} \in \mathbb{R}^{N \times N}$ is a symmetric adjacency matrix, where $A_{ij}$ denotes the edge weight between nodes $v_i$ and $v_j$ such that $A_{ij} > 0$ if $v_i$, $v_j$ are connected and $A_{ij} = 0$ otherwise. The degree of node $v_i$ is defined as $d_i = \sum_{j=1}^NA_{ij}$ and $\mathbf{D} = \text{diag}(d_1, \dots, d_N)$ denotes the degree matrix. The normalized adjacency matrix is defined as $\mathbf{A}_n = \mathbf{D}^{-1/2}\mathbf{A}\mathbf{D}^{-1/2}$. The Laplacian matrix is defined as $\mathbf{L} = \mathbf{D} - \mathbf{A}$ and the normalized Laplacian matrix is $\mathbf{L}_n = \mathbf{I}_N - \mathbf{A}_n$. We define $\mathbf{L}_n = \mathbf{U\Lambda U}^\top$ as the eigen-decomposition of $\mathbf{L}_n$, where $\mathbf{U} \in \mathbb{R}^{N \times N}$ is the matrix of eigenvectors ordered by eigenvalues and $\mathbf{\Lambda} = \text{diag}(\lambda_1, \dots \lambda_N)$ is the diagonal matrix of eigenvalues. 

Let $\mathbf{X} \in \mathbb{R}^{N \times F}$ be the input node feature matrix or graph signals,  $\mathbf{X}_i \in \mathbb{R}^F$ be the features of node $v_i$, and $\mathbf{x}_j \in \mathbb{R}^N$ be the $j$-th signal of all nodes. We measure the smoothness of node features by their total variation w.r.t. a graph $\mathcal{G} = (\mathcal{V}, \mathbf{A})$~\citep{DBLP:journals/tsp/SandryhailaM14}. The total variation of node features is defined as
\begin{equation}\label{e1}
\textbf{TV}(\mathbf{X}) = \textbf{Tr}\left(\mathbf{X}^\top\mathbf{L}_n\mathbf{X}\right),
\end{equation}
\noindent where $\textbf{Tr}(\cdot)$ indicates the trace of a matrix. The smaller $\textbf{TV}(\mathbf{X})$, the smoother are node features $\mathbf{X}$. And we say that $\mathbf{X}$ are smooth if $\textbf{TV}(\mathbf{X})$ is small. 

\subsection{Graph Signal Denoising}\label{app:gsd}
Given a graph $\mathcal{G} = (\mathcal{V}, \mathbf{A})$ with noisy node features $\mathbf{X}$, and assume that the ground-truth node features, denoted as $\hat{\mathbf{X}}$, are smooth, graph signal denoising (GSD)~\citep{DBLP:journals/jstsp/BergerHM17,DBLP:journals/tsp/ChenSMK15} aims to recover smooth graph signals $\hat{\mathbf{X}}$ from noisy input node features $\mathbf{X}$. The GSD problem of a noisy graph can be formulated as an optimization problem that aims to obtain the smoothest node features subject to the effect of noise in the graph.

\subsection{Spectral Graph Convolutions}\label{app:sgc}
SGCNs were mainly inspired by graph signal processing (GSP). In the field of GSP, the \textit{graph Fourier transform}~\citep{hammond2011wavelets,DBLP:journals/tsp/SandryhailaM13} of $\mathbf{x}$ is defined as $\bar{\mathbf{x}} = \mathbf{U}^\top\mathbf{x}$ and the \textit{inverse} graph Fourier transform is defined as $\mathbf{x} = \mathbf{U}\bar{\mathbf{x}}$. Then, the spectral graph convolution of the input signal $\mathbf{x}$ with a filter $\mathbf{g}_{\mathbf{\theta}} = \text{diag}(\mathbf{\theta})$ parameterized by $\mathbf{\theta} \in \mathbb{R}^N$ in the Fourier domain~\citep{DBLP:journals/corr/HenaffBL15} is defined as
\begin{equation}\label{e12}
\mathbf{g}_{\mathbf{\theta}} \star \mathbf{x} = \mathbf{U}\mathbf{g}_{\mathbf{\theta}}\mathbf{U}^\top\mathbf{x},
\end{equation}
\noindent where $\star$ denotes the convolution operation. ChebyNet~\citep{DBLP:conf/nips/DefferrardBV16} uses $K$-order Chebyshev polynomails to approximate the filter $\mathbf{g}_{\mathbf{\theta}}$ and obtains a new convolution:
\begin{equation}\label{e13}
\mathbf{g}_{ChebyNet} \star \mathbf{x} = \sum_{k=0}^{K}\theta_kT_k(\tilde{\mathbf{L}}_n)\mathbf{x},
\end{equation}
\noindent where $T_k(x) = 2xT_{k-1}(x) - T_{k-2}(x)$, $T_0(x) = 1$ and $T_1(x) = x$, $\tilde{\mathbf{L}}_n = \frac{2}{\lambda_{\text{max}}}\mathbf{L}_n - \mathbf{I}_N$, and $\mathbf{\theta}$ is a vector of Chebyshev coefficients. GCN~\citep{DBLP:conf/iclr/KipfW17} introduces a first-order approximation of ChebyNet using $K = 1$ and $\lambda_{\text{max}} = 2$ and the renormalization trick $\mathbf{I}_N + \mathbf{A}_n \rightarrow \tilde{\mathbf{D}}^{-1/2}(\mathbf{I}_N + \mathbf{A})\tilde{\mathbf{D}}^{-1/2} = \tilde{\mathbf{D}}^{-1/2}\tilde{\mathbf{A}}\tilde{\mathbf{D}}^{-1/2}$. The resulting graph convolution becomes
\begin{equation}\label{e14}
\mathbf{g}_{GCN} \star \mathbf{x} = \tilde{\mathbf{A}}_n\mathbf{x},
\end{equation}
\noindent where $\tilde{\mathbf{A}}_n = \tilde{\mathbf{D}}^{-1/2}\tilde{\mathbf{A}}\tilde{\mathbf{D}}^{-1/2}$, $\tilde{\mathbf{D}} = \text{diag}(\tilde{d}_1, \dots, \tilde{d}_N)$, and $\tilde{d}_i = \sum_{j=1}^N\tilde{A}_{ij}$. SGC~\citep{DBLP:conf/icml/WuSZFYW19} successively removes nonlinearities and collapses the weight matrices between $K$ consecutive layers of GCN and further simplifies the convolution as
\begin{equation}\label{e15}
\mathbf{g}_{SGC} \star \mathbf{x} = \tilde{\mathbf{A}}_n^K\mathbf{x}.
\end{equation}
\subsection{Graph Attentions}\label{app:ga}
Unlike SGCN, GANNs were mainly motivated by attention mechanisms in computer vision. GANNs suggest that the contributions of neighboring nodes to the central node are different during the aggregation. Here, we discuss two representative GANN models, GAT~\citep{DBLP:conf/iclr/VelickovicCCRLB18} and AGNN~\citep{DBLP:journals/corr/abs-1803-03735}. GAT utilizes a graph attention mechanism to learn the aggregation strengths between two connected nodes in each layer. Let $\mathbf{H}^{(t)}$ be the $t$-th layer output of a neural network, the graph attention mechanism of GAT is defined as
\begin{equation}\label{e16}
a_{ij}^{(t)} = \text{softmax}\left(f(\mathbf{a}^{\top}[\mathbf{W}\mathbf{H}_i^{(t)} || \mathbf{W}\mathbf{H}_j^{(t)}])_{j \in \mathcal{N}_i \cup \{i\}}\right),
\end{equation}
\noindent where $f(\cdot)$ is the LeakyReLU function, the matrix $\mathbf{W}$ and the vector $\mathbf{a}$ are learnable parameters. Then, GAT aggregates neighborhood information in terms of the learned attention coefficients:
\begin{equation}\label{e17}
\mathbf{H}_i^{(t+1)} = \sigma\left(\sum_{j \in \mathcal{N}_i \cup \{i\}}a_{ij}^{(t)}\mathbf{W}\mathbf{H}_j^{(t)}\right),
\end{equation}
\noindent where $\mathbf{H}^{(0)} = \mathbf{X}$ and $\sigma(\cdot)$ is the activation function. AGNN defines the graph attention mechanism in terms of the cosine similarity of node embeddings between connected nodes:
\begin{equation}\label{e18}
a_{ij}^{(t)} = \text{softmax}\left([\beta\frac{\mathbf{H}_i^{(t)\top} \mathbf{H}_j^{(t)}}{||\mathbf{H}_i^{(t)}||||\mathbf{H}_j^{(t)}||}]_{j \in \mathcal{N}_i \cup \{i\}}\right), 
\end{equation}
\noindent where $\mathbf{H}^{(0)} = \text{ReLU}(\mathbf{X}\mathbf{W})$ and $\beta$ is a learnable parameter. Similarly, AGNN  aggregates neighborhood information by
\begin{equation}\label{e19}
\mathbf{H}_i^{(t+1)} = \sum_{j \in \mathcal{N}_i \cup \{i\}}a_{ij}^{(t)}\mathbf{H}_j^{(t)}.
\end{equation}

\section{Graph Convolutions Work as Denoising and Smoothing Graph Signals}\label{sec:analysis}

In this section, we study the relation between graph signal denoising and SGCNs/GANNs. We show that SGCNs and GANNs are implicitly solving graph signal denoising problems. 

\subsection{Spectral Graph Convolutions Work as Denoising and Smoothing Node Features}\label{section3-1}

We first study the graph signal denoising problem for the case with noisy node features.

\paragraph{Problem 1:} Graph signal denoising for node features.

\begin{figure}[htp]
	\centering
	\includegraphics[width=3.5in]{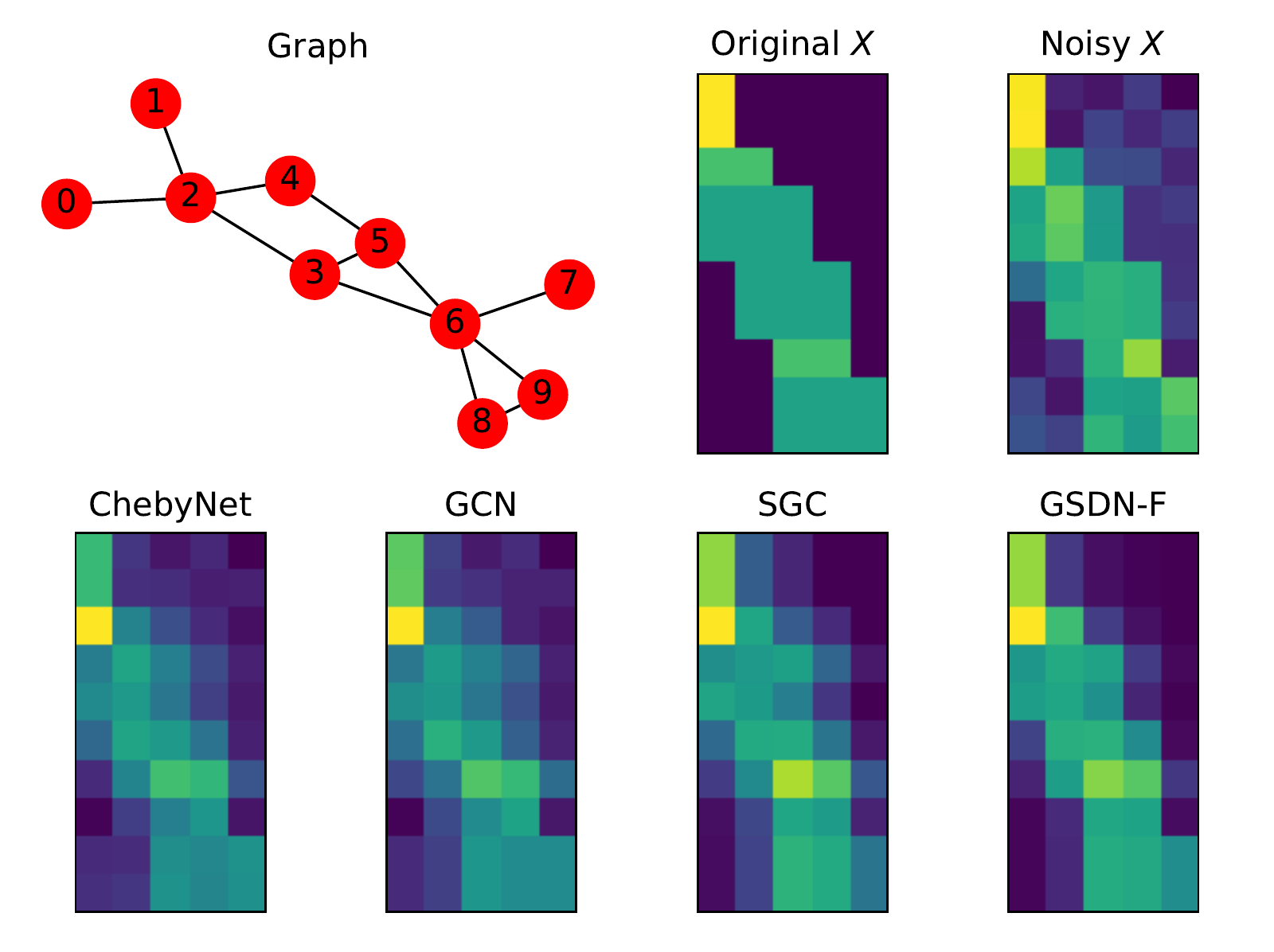}
	\caption{An example of node feature denoising and smoothing by spectral graph convolutions (GSDN-F is to be introduced in Section~\ref{section4-1}). The original node features are disrupted by the Gaussian noise with $\mu = 0$ and $\sigma=0.01$. Best viewed in color}
	\label{figure4}
\end{figure}

We assume that the input node features $\mathbf{X}$ are slightly disrupted with noise and the real (ground-truth) node features, denoted as $\hat{\mathbf{X}}$, are smooth. Formally, we make the following two assumptions.
\begin{assumption}\label{assump1}
	The ground-truth node features $\hat{\mathbf{X}}$ are smooth w.r.t. a graph $\mathcal{G} = (\mathcal{V}, \mathbf{A})$.
\end{assumption}
\begin{assumption}\label{assump2}
	The magnitude of the node feature noise is small.
\end{assumption}
Assumption~\ref{assump1} requires the total variation of $\hat{\mathbf{X}}$ to be small and Assumption~\ref{assump2} implies that $||\hat{\mathbf{X}} - \mathbf{X}||_2$ can be upper-bounded. By Assumptions~\ref{assump1} and~\ref{assump2}, we model Problem 1 as:
\begin{equation}\label{e2}
\begin{aligned}
\hat{\mathbf{X}}^* = \underset{\hat{\mathbf{X}}}{\text{argmin}} \quad \textbf{Tr}\left(\hat{\mathbf{X}}^\top\mathbf{L}_n\hat{\mathbf{X}}\right) \\
\text{s.t.} \quad ||\hat{\mathbf{X}} - \mathbf{X}||_2^2 \leq \epsilon_1,
\end{aligned}
\end{equation}
\noindent where $\epsilon_1 > 0$ controls the noise level. The Lagrangian form of the above problem is $ \mathcal{L}(\hat{\mathbf{X}}, \gamma) = \textbf{Tr}(\hat{\mathbf{X}}^{\top} \mathbf{L}_n\hat{\mathbf{X}}) + \gamma(||\hat{\mathbf{X}} - \mathbf{X}||_2^2 - \epsilon_1) $, where $\gamma > 0$ is the Lagrangian multiplier. Then, we have the following solution by KKT conditions~\citep{gordon2012karush}:
\begin{equation}\label{e3}
\hat{\mathbf{X}}^* = \frac{\gamma}{1 + \gamma}(\mathbf{I}_N - \frac{1}{1 + \gamma}\mathbf{A}_n)^{-1}\mathbf{X}.
\end{equation}
Let $\alpha = \frac{1}{1 + \gamma}$, then $0 < \alpha \leq 1$. The polynomial expansion of $\hat{\mathbf{X}}^*$ is given by
\begin{align}
\hat{\mathbf{X}}^* = {} & (1 - \alpha)(\mathbf{I}_N - \alpha \mathbf{A}_n)^{-1}\mathbf{X} \notag \\
= {} & (1 - \alpha)\left(\mathbf{I}_N + \alpha\mathbf{A}_n + (\alpha\mathbf{A}_n)^2 + \dots\right)\mathbf{X} \notag \\
= {} & (1 - \alpha)\sum_{k=0}^{\infty}(\alpha\mathbf{A}_n)^k\mathbf{X}. \label{e4}
\end{align}
By Eq.\ref{e4}, the following proposition holds.
\begin{proposition}\label{prop1}
	The results of the single layer convolutions of ChebyNet and SGC applied on node feature $\mathbf{X}$ are $K$-order polynomial approximations to the solution $\hat{\mathbf{X}}^*$ of Problem 1, while that of GCN is a first-order polynomial approximation to the solution $\hat{\mathbf{X}}^*$ of Problem 1. 
\end{proposition}
The proof of Proposition~\ref{prop1} is given in Appendix~\ref{app:prop1}. Proposition~\ref{prop1} shows that the graph convolutions of ChebyNet, GCN and SGC are implicitly solving Problem~1. Instead of extracting high-level features, spectral graph convolution operators are simply denoising and smoothing the input node features. We give an example in Figure~\ref{figure4} to show that SGCNs such as GCN and SGC are not extracting high-level features, but simply denoising and smoothing the noisy node features. From the figure, we can see SGCNs such as GCN and SGC are not extracting high-level features, but simply denoising and smoothing the noisy node features.

\begin{figure}[htp]
	\vskip 0.2in
	\begin{center}
		\subfigure[Cora ($\alpha=0.005$)]{
			\includegraphics[width=2in]{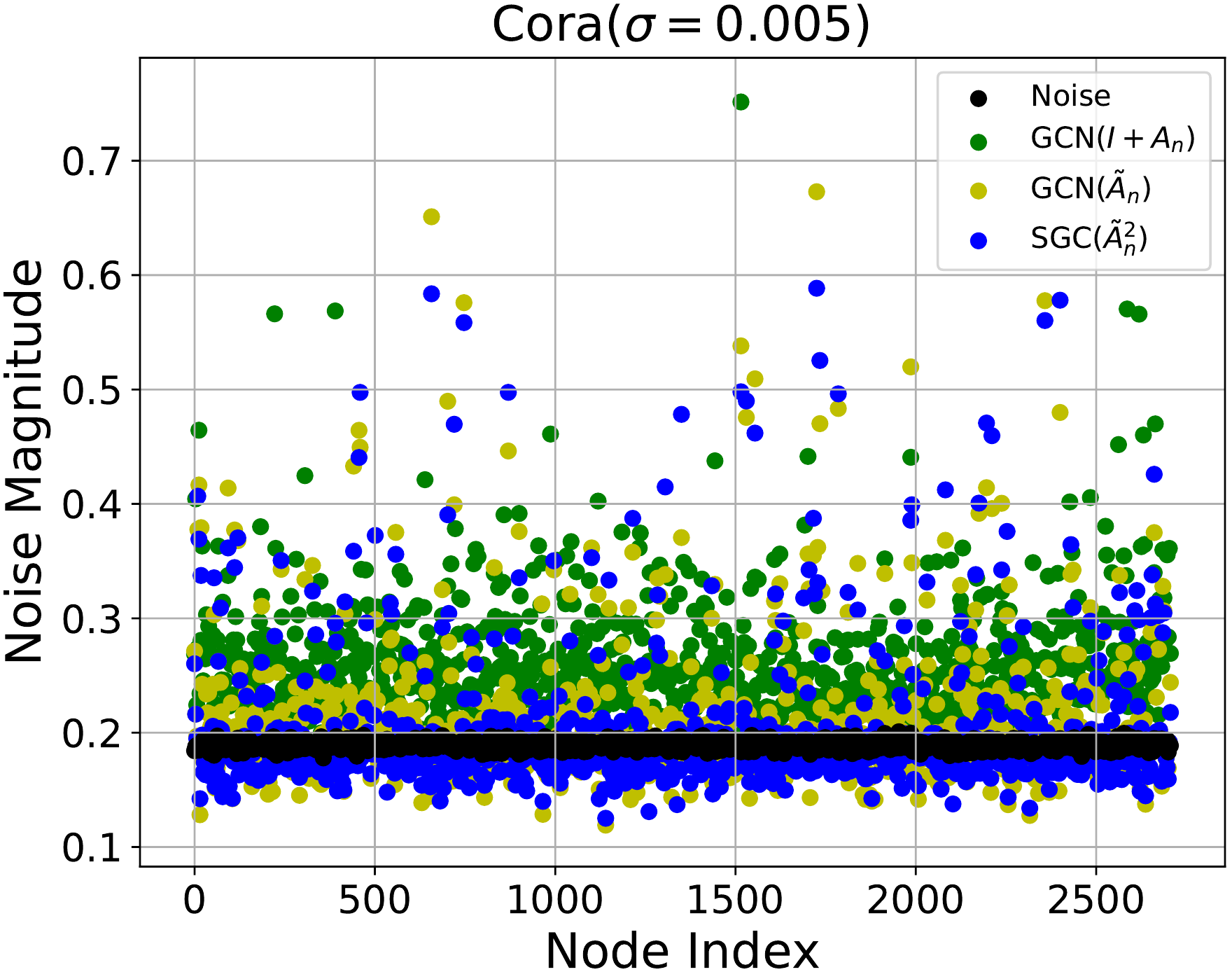}\label{figure5-1}
		}
		\subfigure[CiteSeer ($\alpha=0.005$)]{
			\includegraphics[width=2.05in]{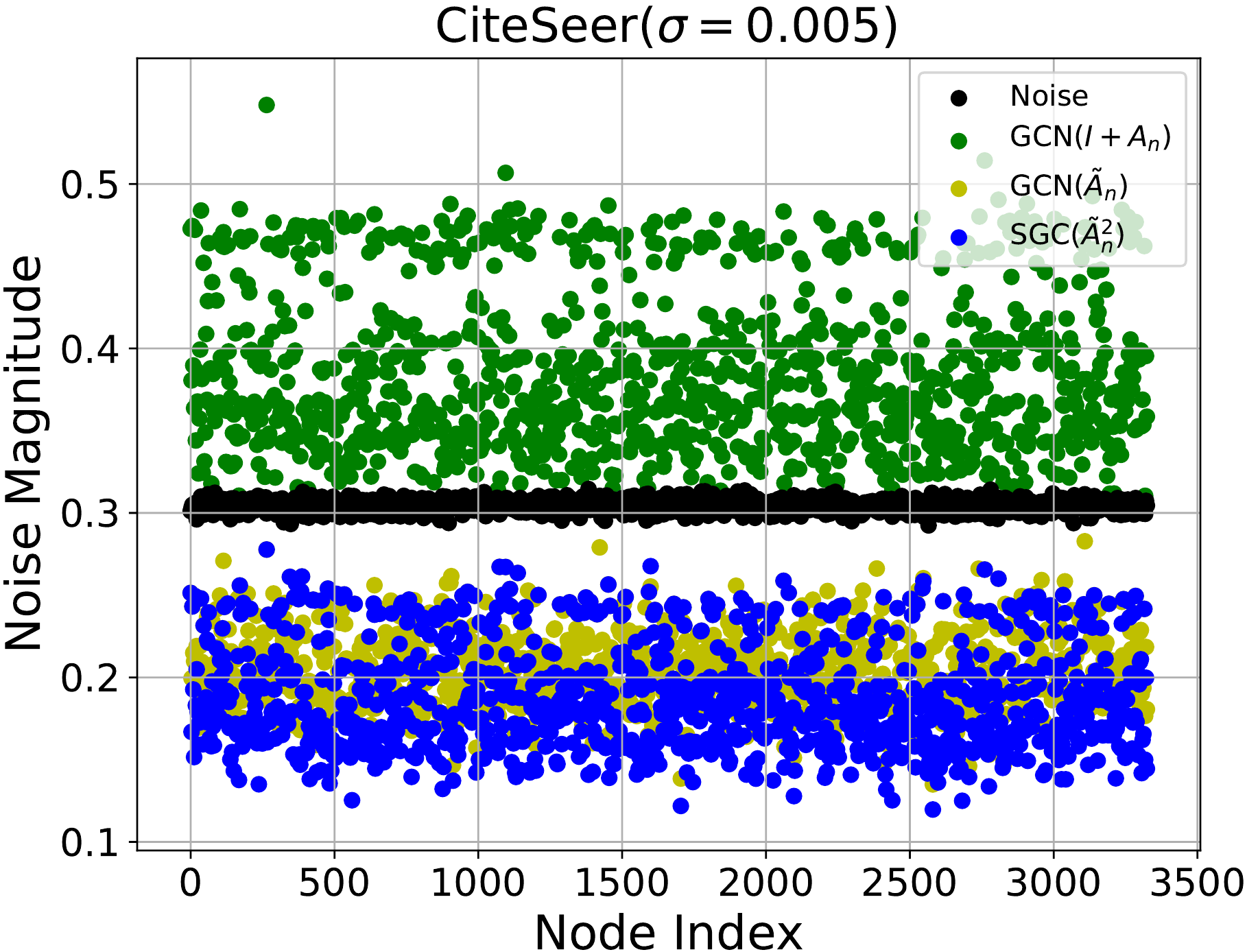}\label{figure5-2}
		} \\
		\subfigure[Cora ($\alpha=0.01$)]{
			\includegraphics[width=2in]{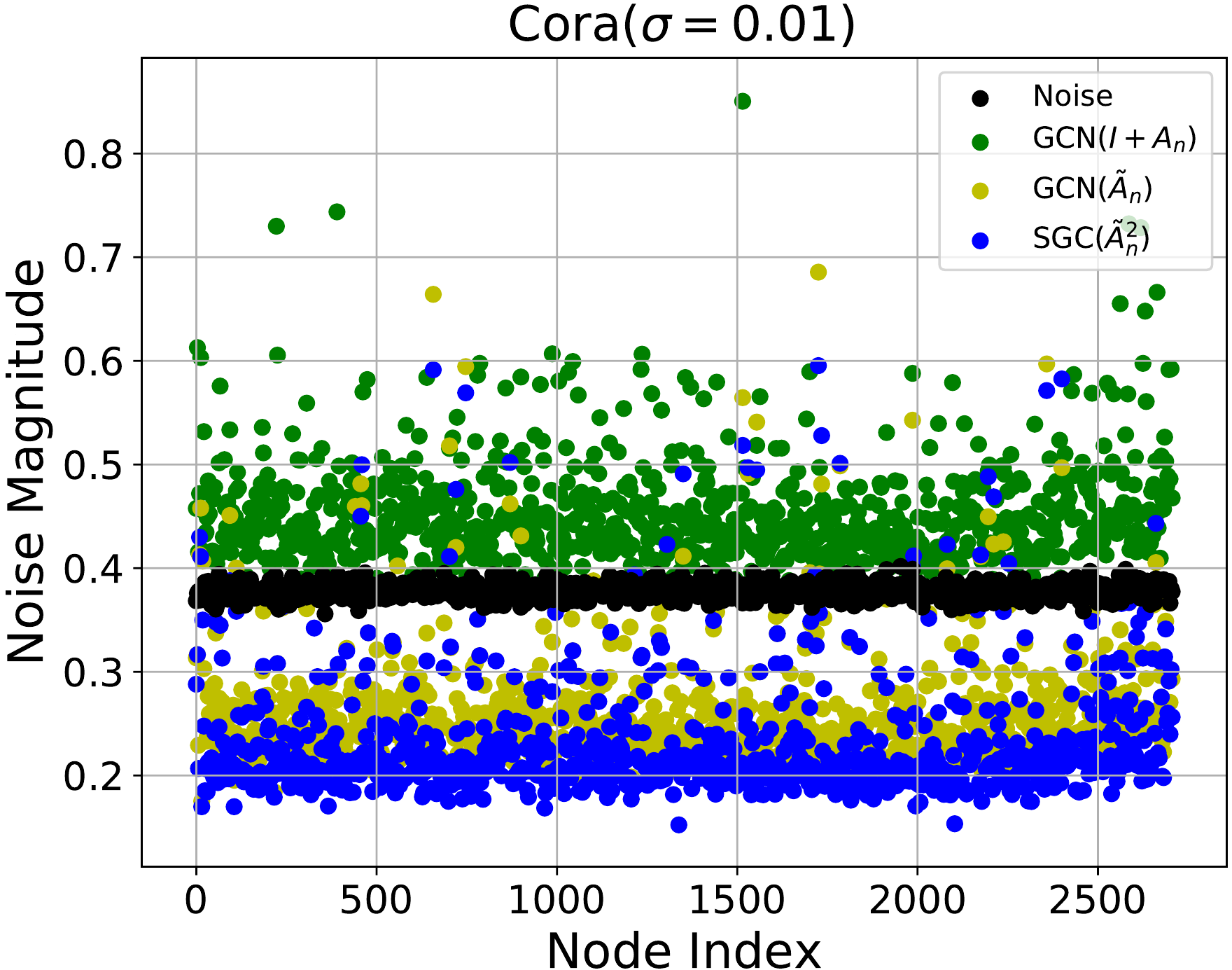}\label{figure5-3}
		}
		\subfigure[CiteSeer ($\alpha=0.01$)]{
			\includegraphics[width=2.05in]{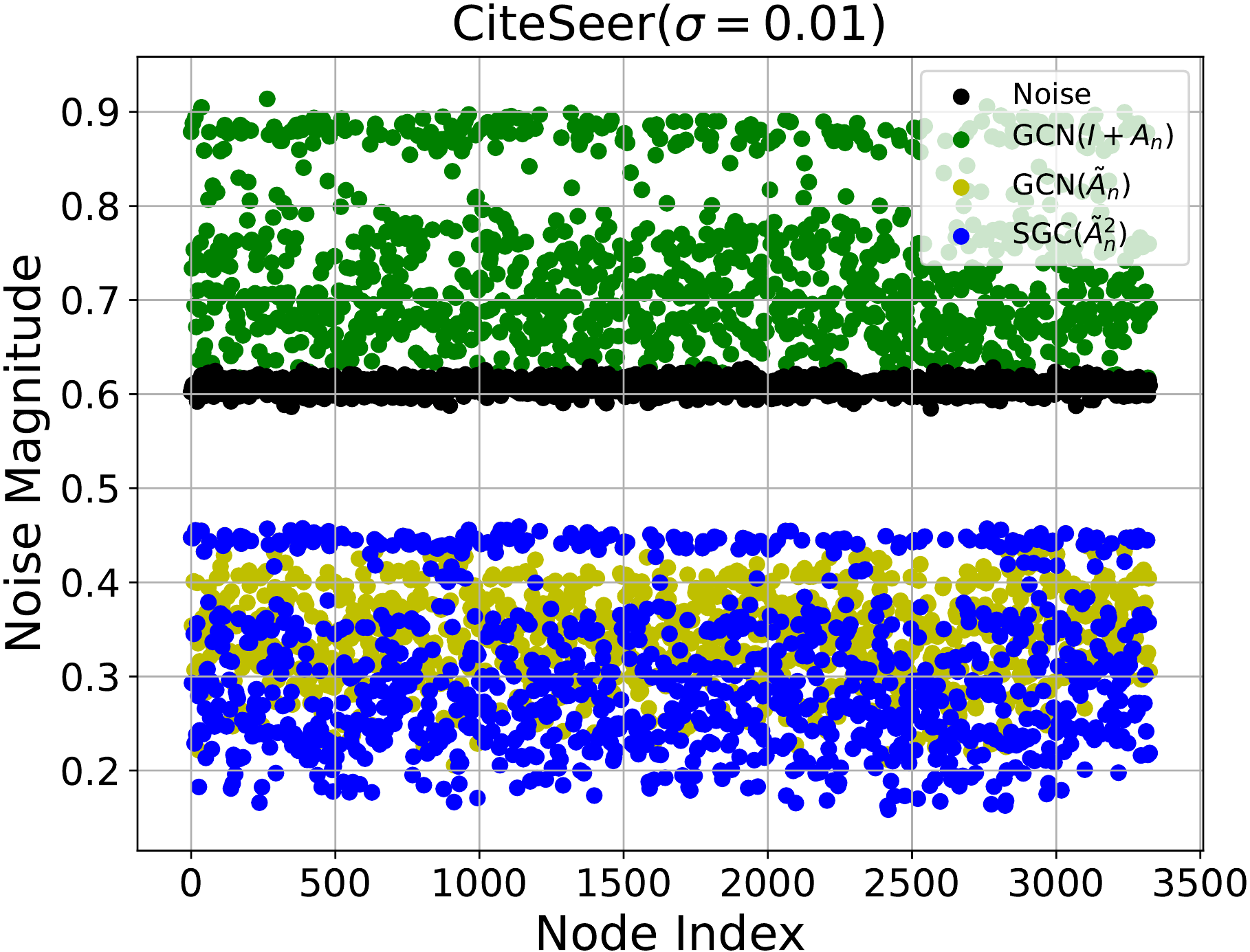}\label{figure5-4}
		}
		\caption{The effect of using the renormalization trick on node feature denoising on graphs with synthetic Gaussian noise ($\mu=0$ and $\sigma=\{0.005, 0.01\}$) on node features (the details of Cora and CiteSeer datasets are to be introduced in Section~\ref{exp:setup})}
		\label{figure5}
	\end{center}
	\vskip -0.2in
\end{figure}

\paragraph{Why Renormalization Trick Works.} We further discuss the effect of the renormalization trick used in GCN. Without the renormalization trick, the result of the graph convolution of GCN on signal $\mathbf{x}$, defined as $\mathbf{g} \star \mathbf{x} = (\mathbf{I}_N + \mathbf{A}_n)\mathbf{x}$, is a first-order polynomial approximation to $\hat{\mathbf{x}}^{*}$ of Problem 1 with $\alpha=1$. As a result of $\alpha=1$, it only intends to smooth node features and overlooks node feature denoising. Therefore, it is easy to be disrupted by noise. The renormalization trick shrinks $\alpha$ into $(0, 1)$ and allows GCN to consider both node feature smoothing and denoising by
\begin{equation}
\mathbf{g}_{GCN} \star \mathbf{x} = \tilde{\mathbf{D}}^{-1/2}(\mathbf{I}_N + \mathbf{A})\tilde{\mathbf{D}}^{-1/2} \approx \tilde{\mathbf{D}}^{-1} + \mathbf{D}_r\mathbf{A}_n, \label{e25}
\end{equation}
\noindent where $\mathbf{D}_r = \text{diag}(d_1 / \tilde{d}_1, \dots, d_N / \tilde{d}_N)$ and $d_i / \tilde{d}_i \leq 1$. Thus, it is more robust to the noise. We empirically validated the effect of the renormalization trick on node feature denoising. The results of node feature denoising on Cora and CiteSeer with Gaussian noise are shown in Figure~\ref{figure5}.

As we can see in Figure \ref{figure5}, the noise in denoised node features obtained by $\mathbf{I} + \mathbf{A}_n$ is not reduced, indicating that $\mathbf{I} + \mathbf{A}_n$ does not denoise node features. It validates that $\mathbf{I} + \mathbf{A}_n$ overlooks node feature denoising by setting $\alpha=1$. However, Figure~\ref{figure5} shows that $\tilde{\mathbf{A}}_n$ and $\tilde{\mathbf{A}}_n^2$ significantly reduce the noise. It demonstrates that graph convolutions with the renormalization trick are capable to denoise node features and validates our conclusion that the renormalization trick works as it shrinks $\alpha$ into $(0, 1)$ and allows GCN to be able to work on node feature denoising.

\subsection{Graph Attentions Work as Denoising Edge Weights}\label{section3-2}
In addition to node features, edge weights may be noisy too in real-world applications. We study the graph signal denoising problem for the case with both noisy node features and noisy edge weights.

\paragraph{Problem 2:} Graph signal denoising for node features and edge weights.

In addition to Assumptions~\ref{assump1} and~\ref{assump2}, we make Assumptions~\ref{assump3} and~\ref{assump4}, which assume that the ground-truth edge weights exactly indicate the smoothness between source nodes and target nodes, and edge weights are slightly disrupted with noise.
\begin{assumption}\label{assump3}
	The ground-truth edge weight $\hat{A}_{ij}$ is inversely proportional to the feature variation between the source node $v_i$ and the target node $v_j$.
\end{assumption}
\begin{assumption}\label{assump4}
	The magnitude of the edge weight noise is small.
\end{assumption}
Assumptions~\ref{assump1} and~\ref{assump3} require that the total variation of $\hat{\mathbf{X}}$ w.r.t. $\mathcal{G}(\mathcal{V}, \hat{\mathbf{A}})$ should be small and Assumption~\ref{assump4} implies that $||\hat{\mathbf{A}}_n - \mathbf{A}_n||_2$ can be upper-bounded. Then, we model Problem 2 below in terms of Assumptions~\ref{assump1}, \ref{assump2}, \ref{assump3} and~\ref{assump4}.
\begin{equation}\label{e5}
\begin{aligned}
\hat{\mathbf{X}}^*, \hat{\mathbf{A}}_n^* = \underset{\hat{\mathbf{X}}, \hat{\mathbf{A}}_n}{\text{argmin}} \quad \textbf{Tr}\left(\hat{\mathbf{X}}^\top\hat{\mathbf{L}}_n\hat{\mathbf{X}}\right) \\
\text{s.t.} 
\quad ||\hat{\mathbf{X}} - \mathbf{X}||_2^2 \leq \epsilon_1, \\
\quad ||\hat{\mathbf{A}}_n - \mathbf{A}_n||_2^2 \leq \epsilon_2, 
\end{aligned}
\end{equation}
\noindent where $\epsilon_1$ and $\epsilon_2$ control the noise level of node features and edge weights, respectively. The Lagrangian form of Problem 2 is $\mathcal{L}(\tilde{\mathbf{X}}, \hat{\mathbf{A}}_n, \gamma, \delta) = \textbf{Tr}(\hat{\mathbf{X}}^{\top} \hat{\mathbf{L}}_n\hat{\mathbf{X}}) + \gamma(||\hat{\mathbf{X}} - \mathbf{X}||_2^2 - \epsilon_1) + \delta(||\hat{\mathbf{A}}_n - \mathbf{A}_n||_2^2 - \epsilon_2)$, where $\gamma > 0$ and $\delta > 0$ are Lagrangian multipliers. Similar to Problem 1, we have the following solution for Problem 2:
\begin{align}
\hat{\mathbf{X}}^* & = \frac{\gamma}{1 + \gamma}(\mathbf{I}_N - \frac{1}{1 + \gamma}\hat{\mathbf{A}}_n^*)^{-1}\mathbf{X}  \label{e6}, \\ 
\hat{\mathbf{A}}_n^* & = \mathbf{A}_n + \sqrt{\epsilon_2}\frac{\hat{\mathbf{X}}^*\hat{\mathbf{X}}^{*\top}}{||\hat{\mathbf{X}}^*||_2^2} \label{e7}.
\end{align}
The forms of graph attentions of GAT and AGNN (as illustrated in Section~\ref{app:ga}) and Eq.\ref{e7} are in the form of calculating the similarity between paired node features. The connection strength of a pair, which is the ground-truth weight $\hat{\mathbf{A}}_{ij}$ in Eq.\ref{e7} or  represented as attention coefficient $a_{ij}$ in GAT and AGNN, depends on the similarity between the features of nodes $v_i$ and $v_j$. 

The differences between them are: (1)~The graph attentions of GAT and AGNN only focus on learning the optimal connection  strength of the connected nodes in a graph. However, Eq.\ref{e7} does not focus only on the connected nodes but any pair of nodes. As a result, it is able to optimize the weights of the existing edges and also create new edges. (2)~There are non-linearities in the  graph attention mechanisms, while Eq.\ref{e7} suggests that a linear form is sufficient to learn the optimal connection strength. Thus, we have Proposition~\ref{prop2}.
\begin{proposition}\label{prop2}
	The attention coefficients of GAT and AGNN can be regarded as the results of denoised weights on the existing edges in a graph.
\end{proposition}
Proposition~\ref{prop2} indicates that the graph attentions of GAT and AGNN are implicitly denoising the weights of the existing edges in a graph.

\section{Graph Signal Denoising Neural Networks}\label{sec:models}
Based on the results of Section~\ref{sec:analysis}, we develop two new GNN models, called GSDN-F and GSDN-EF, as the solutions of Problems~1 and~2, respectively.
\subsection{GSDN-F}\label{section4-1}
For the case where node features are noisy, Section~\ref{section3-1} shows that the graph convolutions of SGCNs work as denoising and smoothing node features $\mathbf{X}$. Their results on $\mathbf{X}$ are polynomial approximations to the solution $\hat{\mathbf{X}}$ of Problem~1. Moreover, they are not able to adjust the balance between denoising and smoothing node features in terms of the noise magnitude in the node features. To this end, based on Eq.\ref{e4}, we design a new graph convolution as
\begin{equation}\label{e8}
\mathbf{g}_{GSDN-F} \star \mathbf{x} = (1 - \alpha) \sum_{k=0}^{K}(\alpha\mathbf{A}_n)^k\mathbf{x},
\end{equation}
\noindent where $\alpha$ controls the balance between smoothing and denoising the node features. We further study the effect of $\alpha$ on the performance of the graph convolution of GSDN-F as follows.

The bias-variance decomposition of the mean square error between the estimator $\hat{\mathbf{x}}^* = \mathbf{g}_{GSDN-F} \star \mathbf{x}$ and the ground-truth $\hat{\mathbf{x}}$ is given as:
\begin{align}
\textbf{MSE}(\hat{\mathbf{x}} | \hat{\mathbf{x}}, \alpha) = {} & \mathbb{E}\left[||\hat{\mathbf{x}}^* - \hat{\mathbf{x}}||_2^2\right] \notag \\
= {} & \mathbb{E}\left[||\hat{\mathbf{x}}^* - \mathbb{E}[\hat{\mathbf{x}}^*]||_2^2\right] + \mathbb{E}\left[||\mathbb{E}[\hat{\mathbf{x}}^*] - \hat{\mathbf{x}}||_2^2\right] \notag \\
= {} & \textbf{Var}(\hat{\mathbf{x}}^* | \alpha) + \textbf{Bias}(\hat{\mathbf{x}}^* | \hat{\mathbf{x}}, \alpha)^2, \label{e9} 
\end{align}
\noindent where $\textbf{MSE}(\cdot)$, $\textbf{Var}(\cdot)$, and $\textbf{Bias}(\cdot)$ represent the mean square error, variance, and bias, respectively. Suppose that the node features are slightly disrupted, i.e., Assumption~\ref{assump2} holds and the noise $\mathbf{z} = \mathbf{x} - \hat{\mathbf{x}}$ follows the normal distribution $\mathbf{z} \sim \textbf{N}(\mathbf{0}, \mathbf{\Sigma})$, we have the following proposition:
\begin{proposition}\label{prop3}
	For $0 < \alpha \leq 1$, $\alpha$ provides a variance and bias trade-off of mean square error between $\hat{\mathbf{x}}^*$ and $\hat{\mathbf{x}}$. Increasing $\alpha$ decreases the variance and increases the bias.
\end{proposition}
\vspace{-1mm}
The proof of Proposition~\ref{prop3} is given in Appendix~\ref{app:prop3}. If the input node features are severely disrupted, then Assumption~\ref{assump2} no longer holds, implying that the noise can be lower-bounded, i.e., $||\hat{\mathbf{X}} - \mathbf{X}||_2^2 \ge \epsilon_1$. Then we can obtain $\alpha=\frac{1}{1-\gamma}$ in terms of the methods of Lagrangian multipliers and KKT conditions. Thus, we suggest to choose $\alpha > 1$ in this case. We also conduct experiments on node classification to study the parameter sensitivity of $\alpha$. The results are given in Appendix~\ref{app:sensitivity}.

\subsection{GSDN-EF}\label{section4-2}
For the case when both node features and edges are noisy, we design the graph convolution of GSDN-EF in terms of the solutions $\hat{\mathbf{A}}$ and $\hat{\mathbf{X}}$ of Problem 2. The new graph convolution scheme is designed as first denoising the adjacency matrix to obtain $\hat{\mathbf{A}}$ following Eq.\ref{e10}, and then aggregating the input signal $\mathbf{x}$ by Eq.\ref{e11} based on $\hat{\mathbf{A}}$:

\vspace{-5mm}

\begin{align}
\hat{\mathbf{A}} & = \mathbf{A}_n + \beta\frac{\mathbf{X}\mathbf{X}^\top}{||\mathbf{X}||_2^2}, \label{e10} \\
\mathbf{g}_{GSDN-EF} \star \mathbf{x} & = (1 - \alpha) \sum_{k=0}^{K}(\alpha\hat{\mathbf{A}}_n)^k\mathbf{x}, \label{e11}
\end{align}
\noindent where $\beta$ is a learnable parameter, $\hat{\mathbf{A}}_n = \hat{\mathbf{D}}^{-1/2}\hat{\mathbf{A}}\hat{\mathbf{D}}^{-1/2}$, $\hat{\mathbf{D}} = \text{diag}(\hat{d}_1, \dots, \hat{d}_N)$ and $\hat{d}_i = \sum_{j=1}^N\hat{A}_{ij}$. While GAT and AGNN only focus on optimizing the connection strength between connected nodes (i.e., edges), GSDN-EF considers any pair of nodes and does not only optimize the weights of the existing edges but is also able to create new edges. GSDN-EF can be effective for the cases where we miss some edges in a graph during data collection or graph construction.

\section{Experiments and Discussion}\label{sec:exp}

We validate our theoretical findings by experiments on benchmark graph datasets with/without noise. We conducted experiments for node feature denoising and smoothing, as well as for semi-supervised node classification.

\subsection{Datasets and Experimental Setup}\label{exp:setup}

Here we present the details of the datasets and the experimental setup.

\paragraph{Datasets.} We used Cora, CiteSeer, and Pubmed citation networks~\citep{DBLP:journals/aim/SenNBGGE08} in the  transductive learning task and the PPI protein-protein interaction network~\citep{DBLP:journals/bioinformatics/ZitnikL17} in the inductive learning task. We also conducted experiments by adding noise to node features or edges in the Cora and CiteSeer datasets to evaluate the performance of GNNs on graphs with noise. Some information about the datasets are listed in Table~\ref{tabel3}.

\begin{table}[htp]
	\caption{Datasets}\label{tabel3}
	\centering
	\begin{tabular}{lccccr}
		\toprule
		Dataset & \#Nodes & \#Edges & Train/Dev/Test \\
		\midrule
		Cora & 2,708 & 5,429 & 140/500/1,000 \\
		CiteSeer & 3,327 & 4,723 & 120/500/1,000 \\
		Pubmed & 19,717 & 44,338 & 60/500/1,000 \\
		\midrule
		PPI & 56,944 & 818,716 & 44,906/6,514/5,524 \\
		\bottomrule
	\end{tabular}
\end{table}

\paragraph{Baselines.} For the baselines of spectral approaches of GNNs, we compared our models with ChebyNet~\citep{DBLP:conf/nips/DefferrardBV16}, GCN~\citep{DBLP:conf/iclr/KipfW17}, and SGC~\citep{DBLP:conf/icml/WuSZFYW19}. For non-spectral approaches, we chose GraphSage~\citep{DBLP:conf/nips/HamiltonYL17}, GAT~\citep{DBLP:conf/iclr/VelickovicCCRLB18}, and AGNN~\citep{DBLP:journals/corr/abs-1803-03735}. We used the released implementations of these baselines in PyTorch Geometric~\citep{Fey/Lenssen/2019}.

\paragraph{Parameter Setting.} We give the parameter setting for our models and baselines in our experiments here. For GSDN-F and GSDN-EF, we set the learning rate as $0.02$, $L2$ regularization weight as $5 \times 10^{-4}$ and the polynomial degree $K$ as 4. In transductive learning tasks, we set the number of hidden units as $16$, the number of layers as $2$. We also use two settings for $\alpha$, 0.6 and 1.2, for both GSDN-F and GSDN-EF and denote the models with different $\alpha$ as \textbf{GSDN-F-0.6}, \textbf{GSDN-F-1.2}, \textbf{GSDN-EF-0.6}, and \textbf{GSDN-EF-1.2}, respectively. For the inductive learning task, we set hidden number as 256, the number of layers as 3 and $\alpha$ as 0.6. For ChebyNet and SGC, we used two settings for $K$, 2 and 4, represented as ChebyNet-2, ChebyNet-4, SGC-2, and SGC-4. Other hyperparameters for ChebyNet, SGC, and other baselines were set by following the settings in \citep{DBLP:conf/iclr/KipfW17} and \citep{DBLP:conf/iclr/VelickovicCCRLB18}.

\subsection{Node Feature Denoising and Smoothing}\label{exp:denoising-smoothing}

We first normalized the node features and then added Gaussian noise with mean $\mu=0$ and standard deviation $\sigma=\{0.001, 0.005, 0.01\}$ to the original node features (denoted as $\hat{\mathbf{X}}^*$) in the Cora and CiteSeer graphs. We studied the performance of spectral graph convolutions on node feature denoising and smoothing in this experiment.

\begin{figure}[!t]
	\centering
	\subfigure[Cora ($\sigma=0.001$)]{
		\includegraphics[width=1.6in]{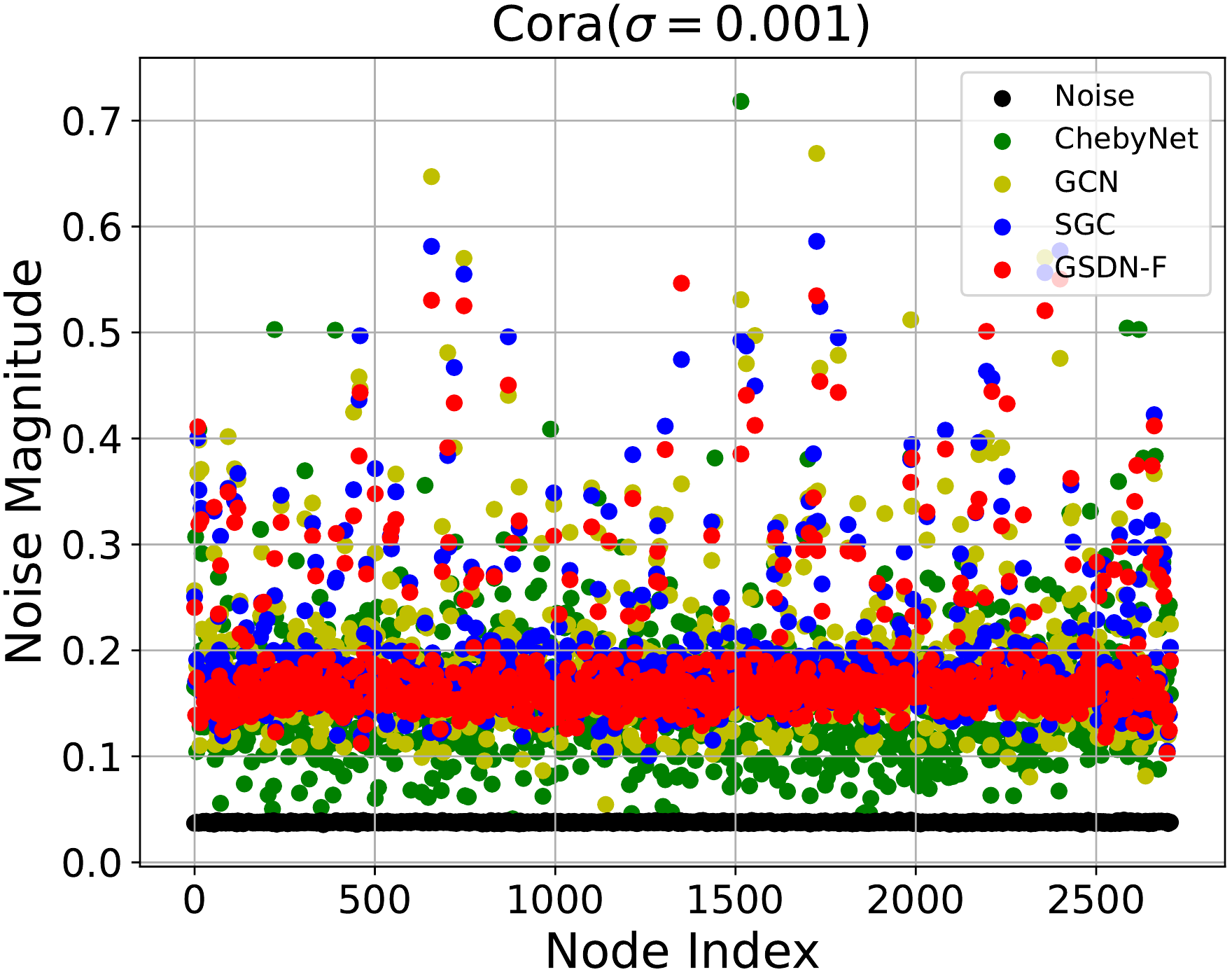}
	}
	\subfigure[Cora ($\sigma=0.005$)]{
		\includegraphics[width=1.6in]{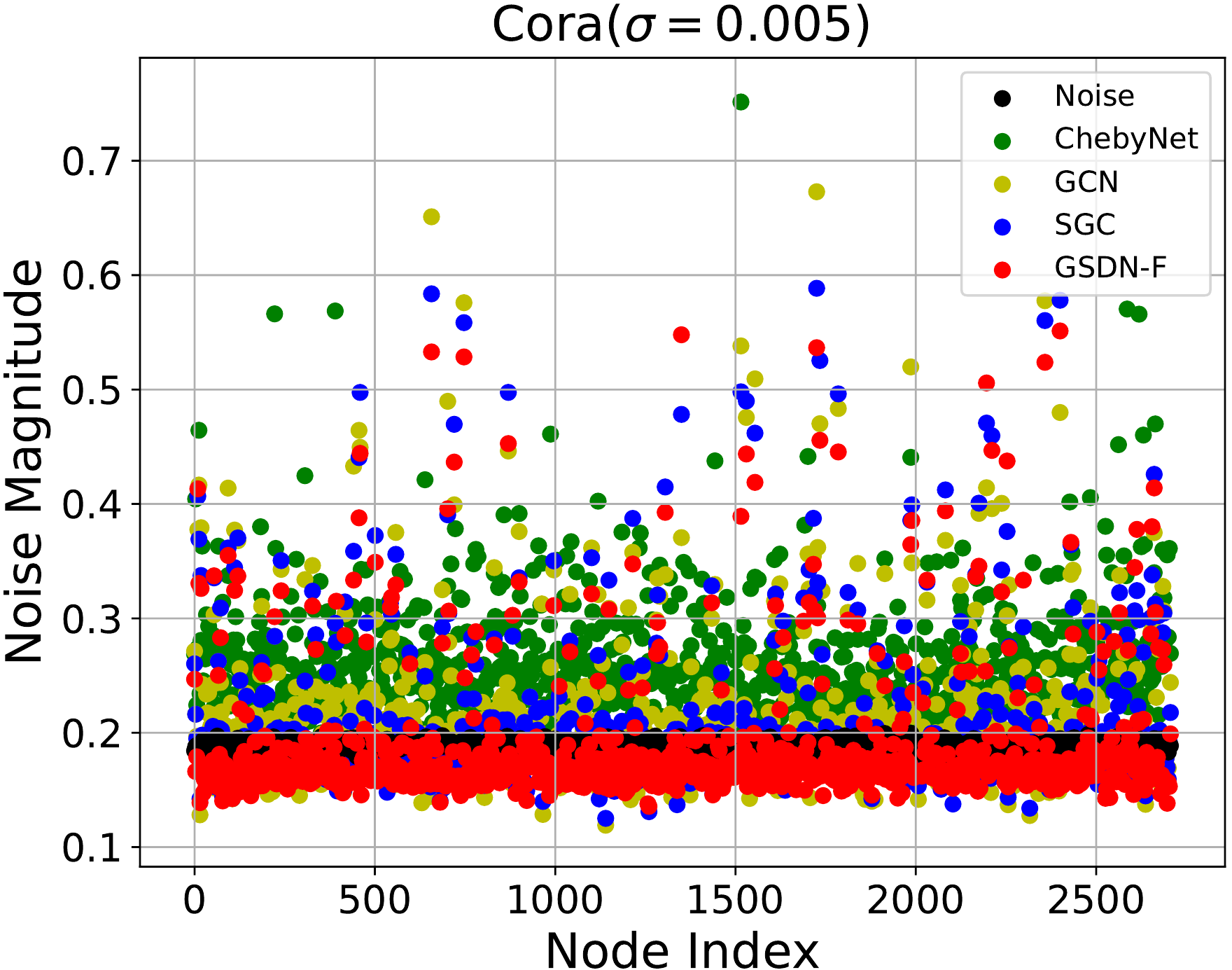}
	}
	\subfigure[Cora ($\sigma=0.01$)]{
		\includegraphics[width=1.62in]{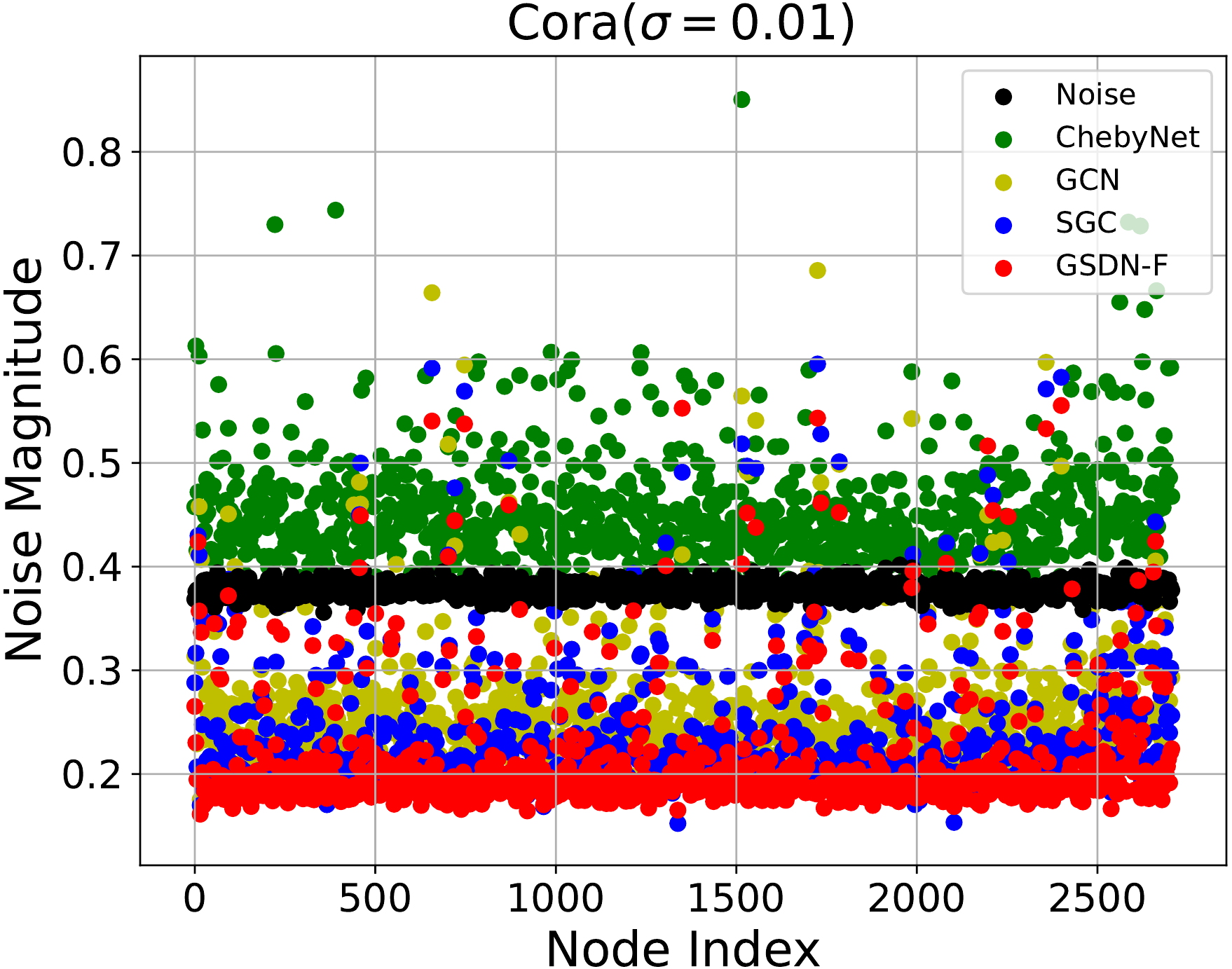}
	}
	\subfigure[CiteSeer ($\sigma=0.001$)]{
		\includegraphics[width=1.67in]{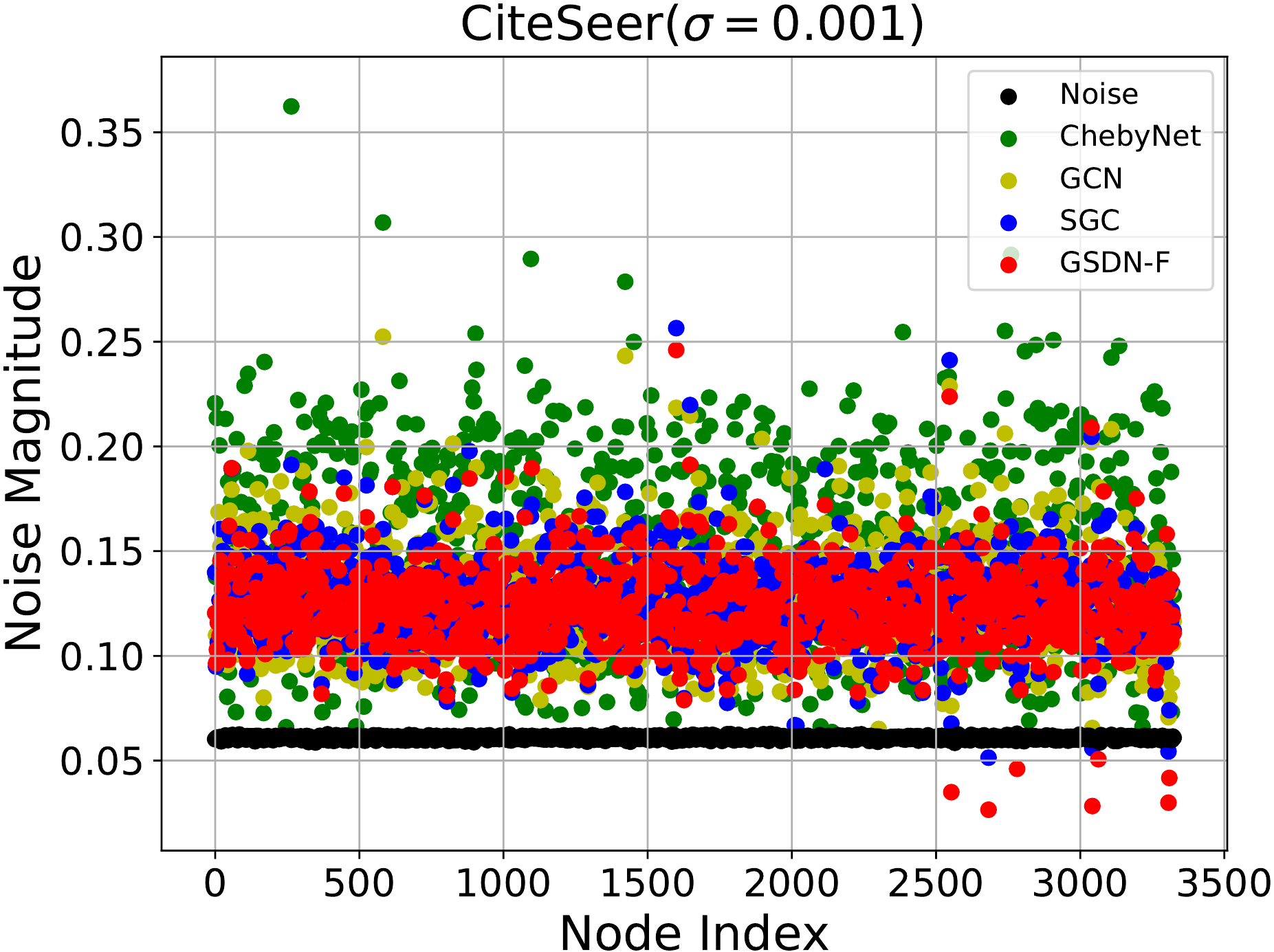}
	}
	\subfigure[CiteSeer ($\sigma=0.005$)]{
		\includegraphics[width=1.67in]{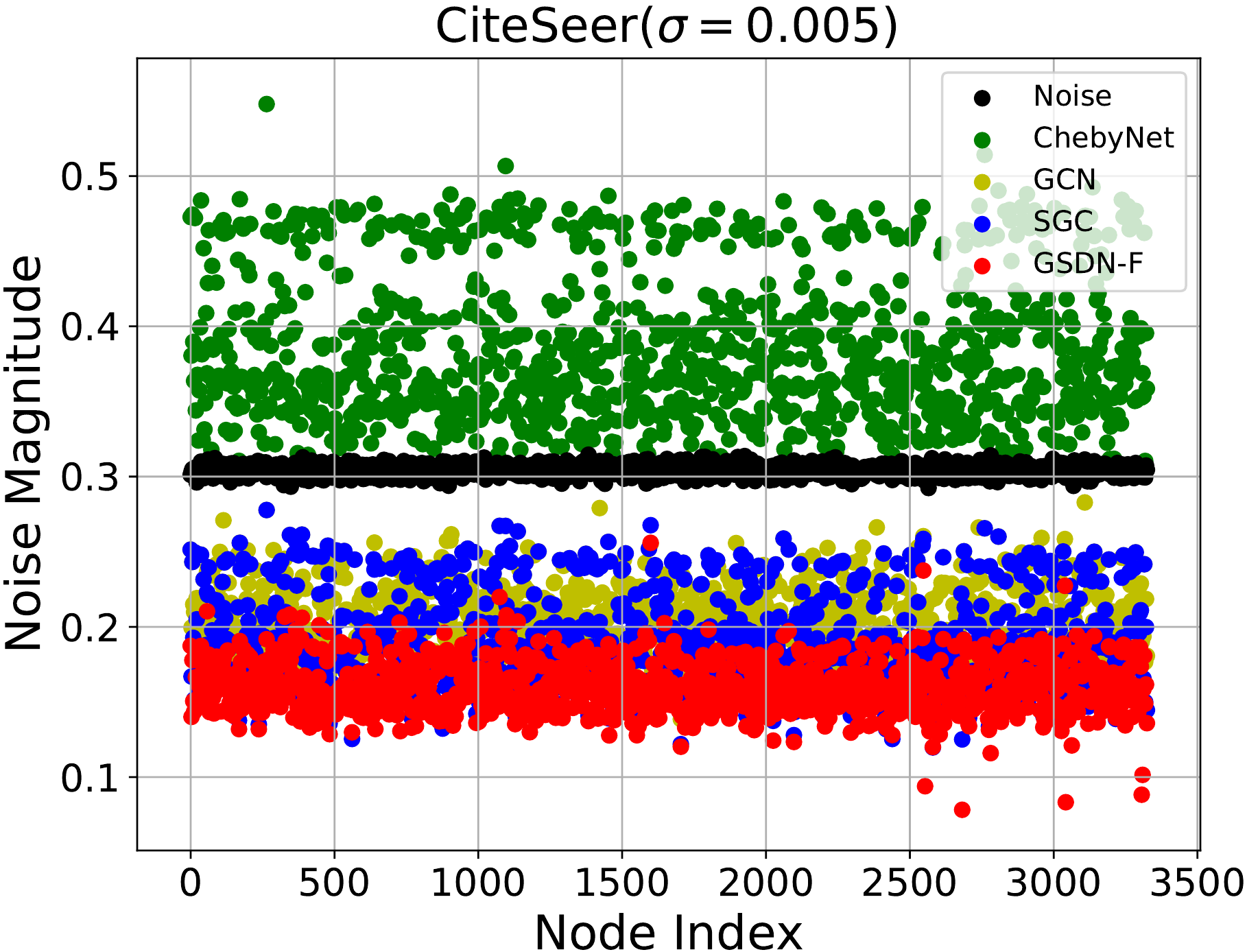}
	}
	\subfigure[CiteSeer ($\sigma=0.01$)]{
		\includegraphics[width=1.65in]{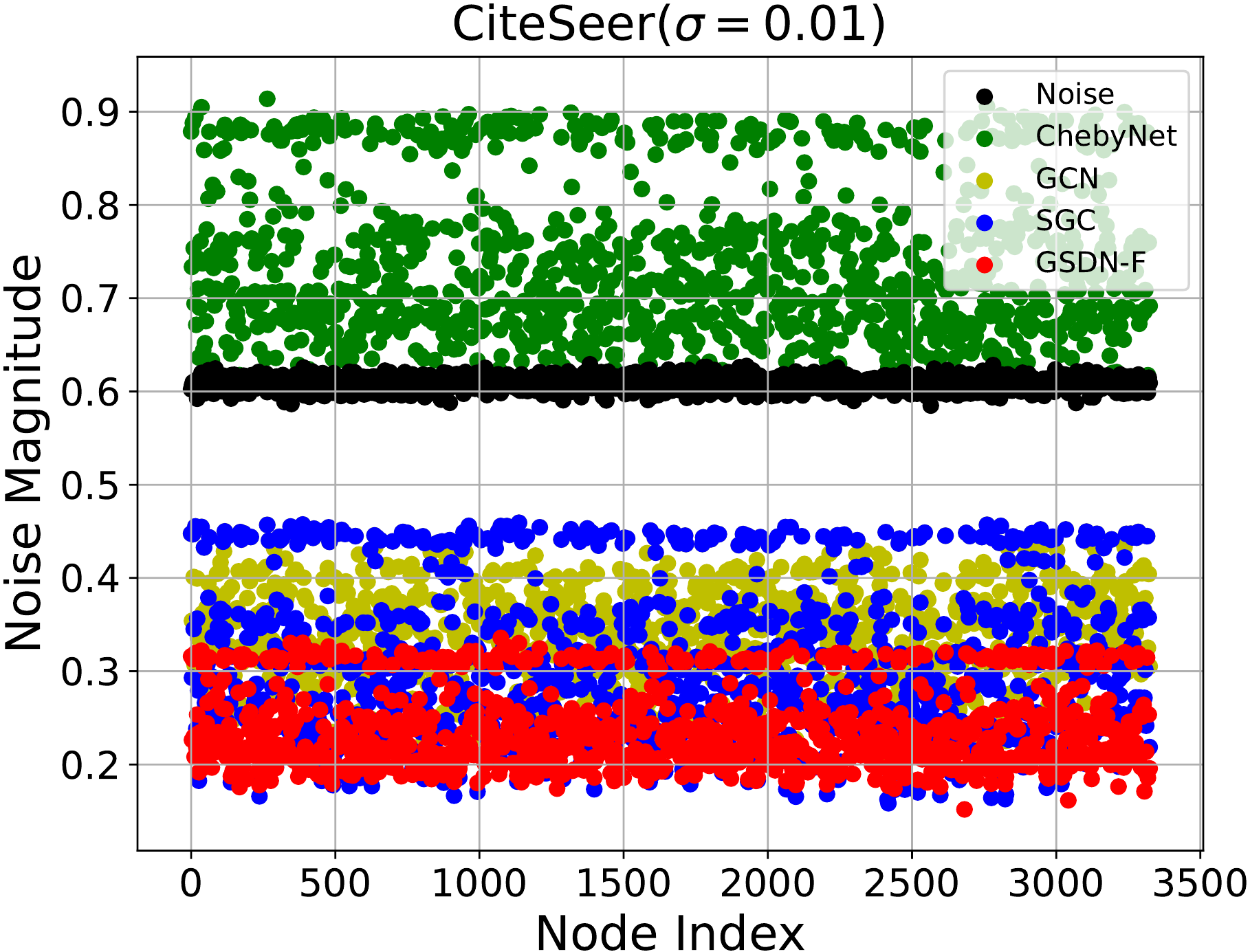}
	}
	\caption{Results of node feature denoising by graph convolutions (best viewed in color)}
	\label{figure1}
\end{figure}

\paragraph{Node feature denoising.} We used a spectral graph convolution $\mathbf{g}$ to denoise node features $\mathbf{X}$ by $\hat{\mathbf{X}} = \mathbf{g} \star \mathbf{X}$, where $\hat{\mathbf{X}}$ is the denoised features. The noise magnitude of each node, measured by $||\hat{\mathbf{X}}_i^* - \hat{\mathbf{X}}_i||_2$, is reported in Figure~\ref{figure1}. The results show that when there is little noise in node features ($\sigma=0.001$), all spectral graph convolutions do not reduce noise. When $\sigma \geq 0.005$, the graph convolutions of SGC, GCN, and GSDN-F ($\alpha=0.6$) are denoising node features. In all the cases, GSDN-F obtains the best overall performance, i.e., the lowest noise magnitude, while ChebyNet is not denoising the node features.

\begin{figure}[!t]
	\centering
	\subfigure[Cora]{
		\includegraphics[width=2in]{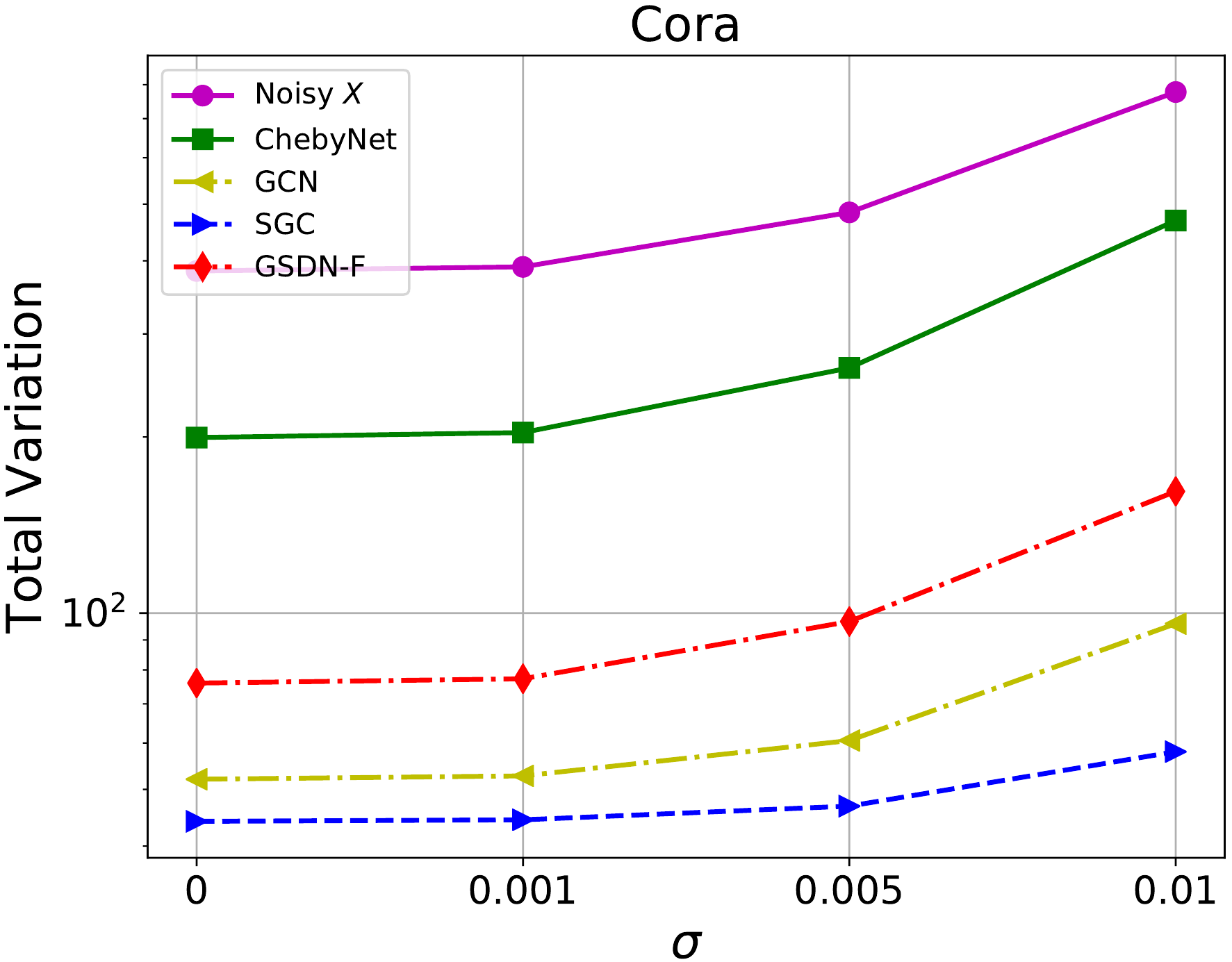}
	}
\hspace{20mm}
	\subfigure[CiteSeer]{
		\includegraphics[width=2in]{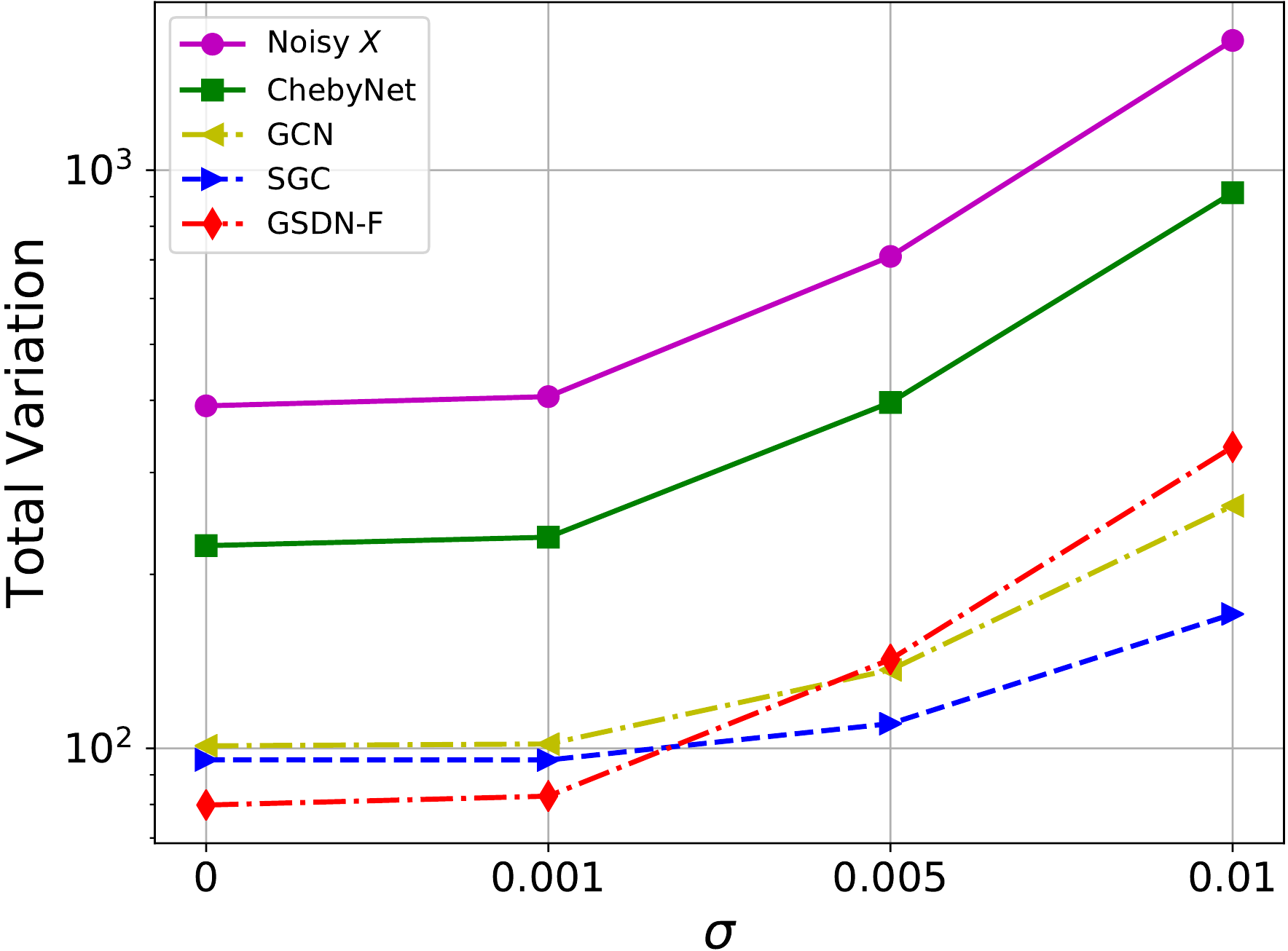}
	}
	\caption{Total variation (the smaller, the smoother) of node features (best viewed in color)}
	\label{figure2}
\end{figure}

\paragraph{Node feature smoothing.} We measured the smoothness of the denoised node features by their total variation as defined in Eq.\ref{e1}. Figure~\ref{figure2} shows that the results of all spectral graph convolutions are smoother than the original noisy features $\mathbf{X}$, indicating that they all smooth the input node features. When there is less noise ($\sigma \leq 0.005$) in the node features, the results of GCN, SGC, and GSDN-F ($\alpha=0.6$) are close to each other. However, when there is more noise (e.g., $\sigma=0.01$), the results of GCN and SGC are smoother than that of GSDN-F. In all cases, the results of ChebyNet are much less smooth than those of the others.

\paragraph{Analysis.} The results of Figures~\ref{figure1} and~\ref{figure2} show that the graph convolutions of GCN, SGC, and GSDN-F are denoising and smoothing node features. When there is little noise (e.g., $\sigma=0.001$) in the node features, all of the methods are simply smoothing node features. However, when there is more noise (e.g., $\sigma \geq 0.005$), GSDN-F conducts more denoising, while GCN and SGC conduct more smoothing. The results show that GSDN-F is more likely to denoise node features, while GCN and SGC are more committed to smoothing in the cases with more noise. 


\subsection{Semi-Supervised Node Classification}\label{exp:classification}

To explore how GNNs benefit from node feature and/or edge weight denoising and node feature smoothing, we conducted experiments for semi-supervised node classification on graphs with/without noise in node features and edges.

\paragraph{Results on graphs without noise.} Table~\ref{tabel1} reports the classification accuracy of transductive learning on Cora, CiteSeer, and Pubmed. The results show that the performance of GCN, SGC, and GSDN-F is close to each other. Their similar node classification accuracy can be explained by their similar performance on node feature denoising and smoothing on graphs with little noise (i.e., $\sigma=0.001$) as reported in Figures~\ref{figure1} and~\ref{figure2}. Table~\ref{tabel2} reports the Micro-F1 score of inductive learning on the PPI dataset. As GSDN-EF ran out of memory on PPI, we used a sparse version, GSDN-EF(Sparse), which only denoises the original edges in a graph (but does not create new edges). The results in Table~\ref{tabel2}, and also Table~\ref{tabel1}, show the performance of GSDN-EF is comparable with that of GAT and AGNN, while it significantly outperforms the spectral approaches (i.e., GCN, SGC, ChebyNet).  We remark that our paper focuses on theoretical understandings of GNN models, while the experiments serve more as a validation of our theoretical findings, i.e., the performance of SGCNs on node classification benefits from their ability of node feature denoising and smoothing, while GANNs benefits from their ability of edge weight denoising. Thus,  on graphs without noise, our models only have comparable performance with existing models because of their similar performance on denoising and smoothing. 

\begin{table}[htp]
	\centering
	\caption{Node classification accuracy (\%) averaged over 20 runs on citation networks}\label{tabel1}
	{\small
	\textbf{Transductive} \\
	\begin{tabular}{lcccr}
		\toprule
		Method & Cora & CiteSeer & Pubmed \\
		\midrule
		GCN & 81.3 $\pm$ 0.7 & 70.8 $\pm$ 0.9 & 78.6 $\pm$ 0.7 \\
		ChebyNet-2 & 79.2 $\pm$ 0.8 & 70.1 $\pm$ 0.8 & 78.0 $\pm$ 0.6 \\
		ChebyNet-4 & 80.1 $\pm$ 0.9 & 70.0 $\pm$ 1.2 & 73.3 $\pm$ 2.1 \\
		SGC-2 & 79.6 $\pm$ 0.6 & 72.0 $\pm$ 0.8 & 77.5 $\pm$ 0.2 \\
		SGC-4 & 81.0 $\pm$ 0.6 & 72.8 $\pm$ 0.4 & 75.1 $\pm$ 0.2 \\
		GraphSage & 81.5 $\pm$ 0.8 & 70.3 $\pm$ 0.7 & 78.6 $\pm$ 0.4 \\
		GAT & 82.3 $\pm$ 0.7 & 71.3 $\pm$ 0.8 & 78.1 $\pm$ 0.6 \\
		AGNN & 82.0 $\pm$ 0.6 & 70.9 $\pm$ 0.9 & 79.2 $\pm$ 0.5 \\
		\midrule
		GSDN-F-0.6 & 81.5 $\pm$ 0.8 & 70.6 $\pm$ 1.2 & 79.0 $\pm$ 0.8 \\
		GSDN-F-1.2 & 81.0 $\pm$ 0.7 & 70.0 $\pm$ 0.6 & 78.7 $\pm$ 0.5 \\
		GSDN-EF-0.6 & 82.6 $\pm$ 0.7 & 71.1 $\pm$ 1.0 & 78.9 $\pm$ 0.4 \\
		GSDN-EF-1.2 & 81.9 $\pm$ 0.7 & 70.0 $\pm$ 0.9 & 78.1 $\pm$ 1.7 \\
		\bottomrule
	\end{tabular}
}
\end{table}
\begin{table}[htp]
	\centering
	\caption{Micro-F1 score (\%) averaged over 10 runs on the PPI dataset}\label{tabel2}
	{\small
	\textbf{Inductive} \\
	\begin{tabular}{lcr}
		\toprule
		Method & PPI \\
		\midrule
		GCN & 68.7 $\pm$ 1.7\\
		AGNN & 84.8 $\pm$ 1.3\\
		GAT & 97.1 $\pm$ 0.7\\
		\midrule
		GSDN-F & 77.2 $\pm$ 2.6\\
		GSDN-EF(sparse) & 95.7 $\pm$ 0.8\\
		\bottomrule
	\end{tabular}
}
\end{table}

\begin{figure}[t]
	\centering
	\subfigure[Cora w/ node feature noise\label{figure3-1}]{
		\includegraphics[width=1.62in]{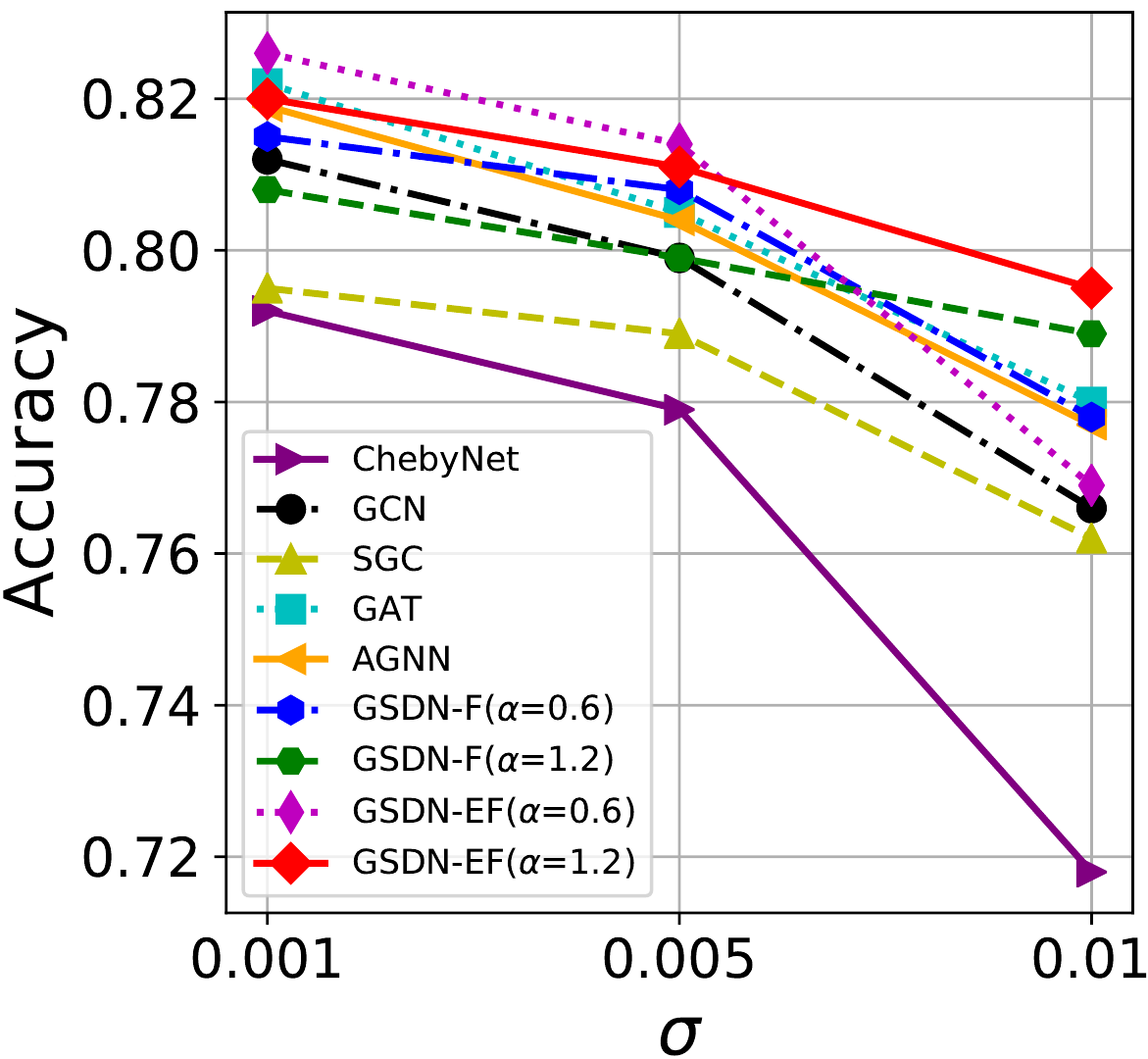}
	}
	\subfigure[Cora w/ edge noise\label{figure3-2}]{
		\includegraphics[width=1.62in]{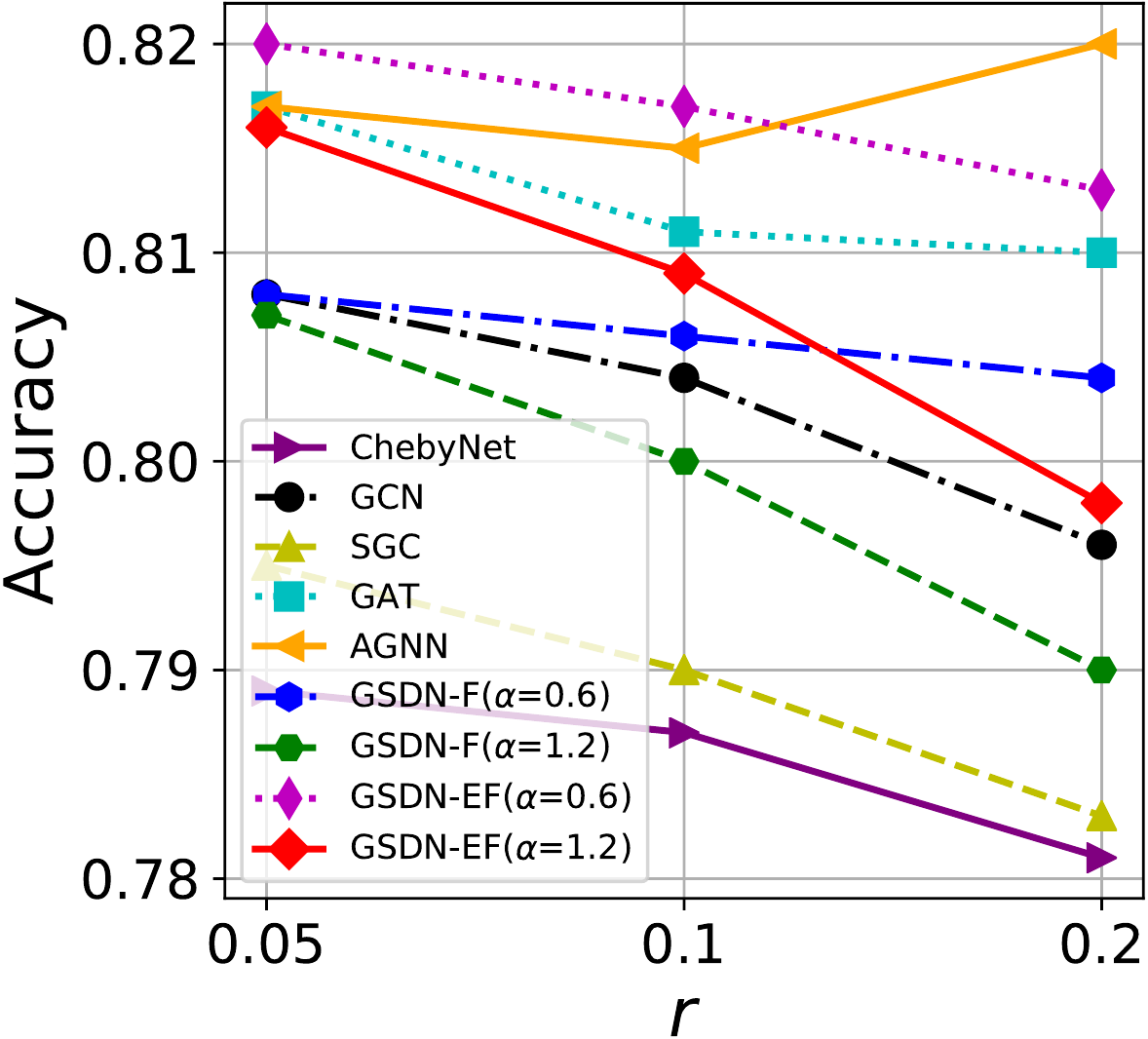}
	}
	\subfigure[Cora w/ node feature noise and edge noise\label{figure3-3}]{
		\includegraphics[width=1.63in]{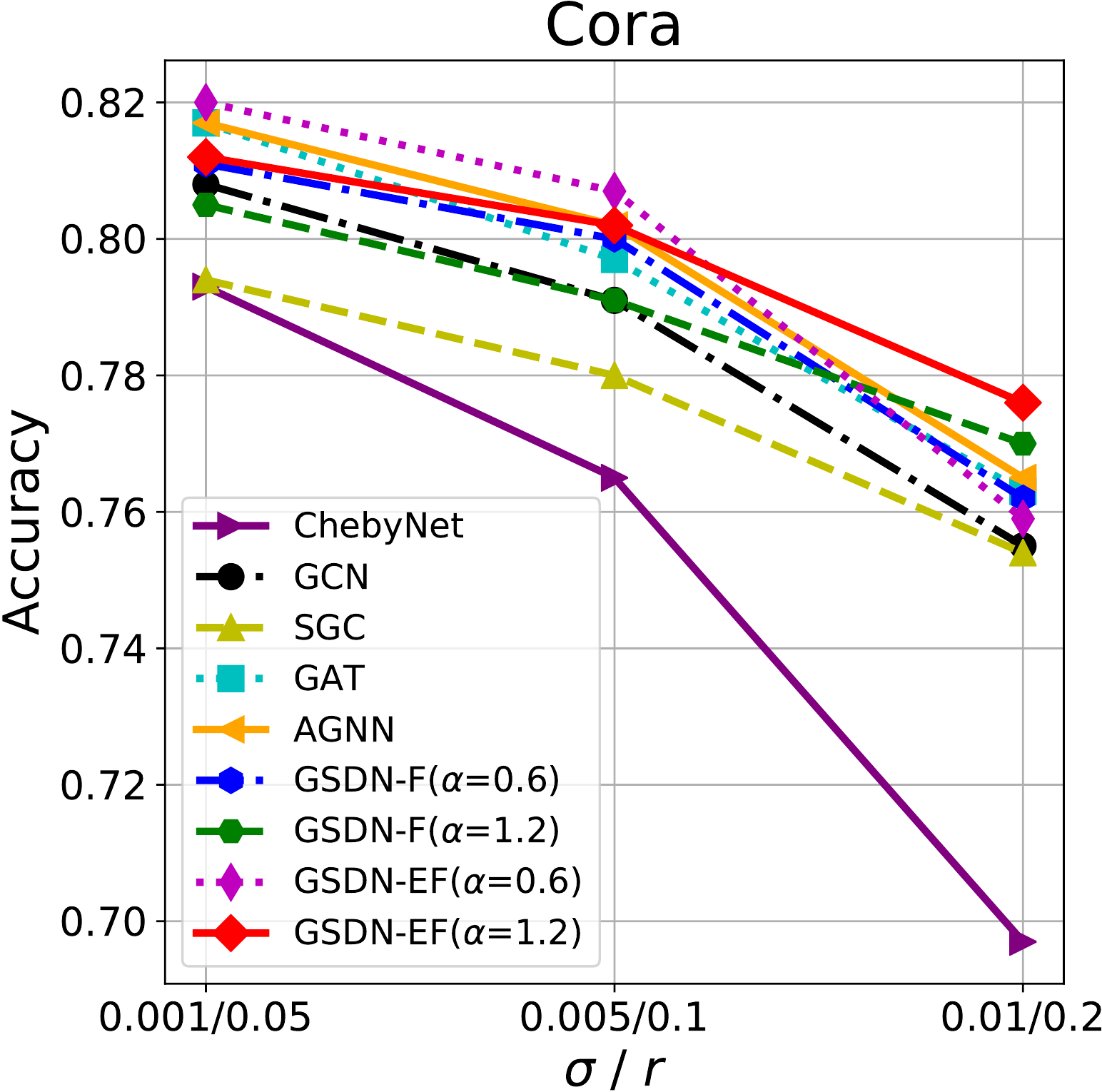}
	}
	\subfigure[CiteSeer w/ node feature noise\label{figure3-4}]{
		\includegraphics[width=1.62in]{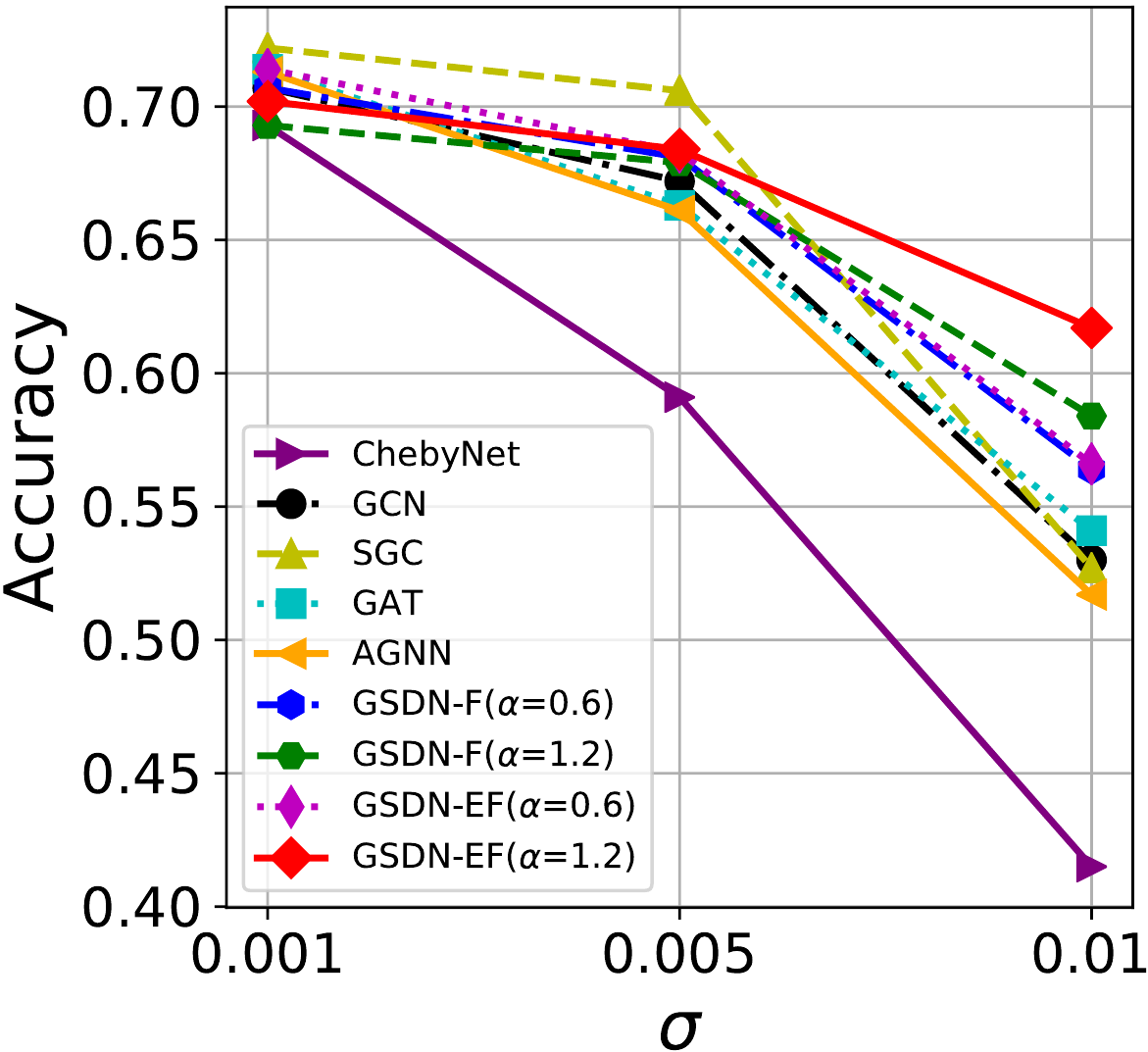}
	}
	\subfigure[CiteSeer w/ edge noise\label{figure3-5}]{
		\includegraphics[width=1.62in]{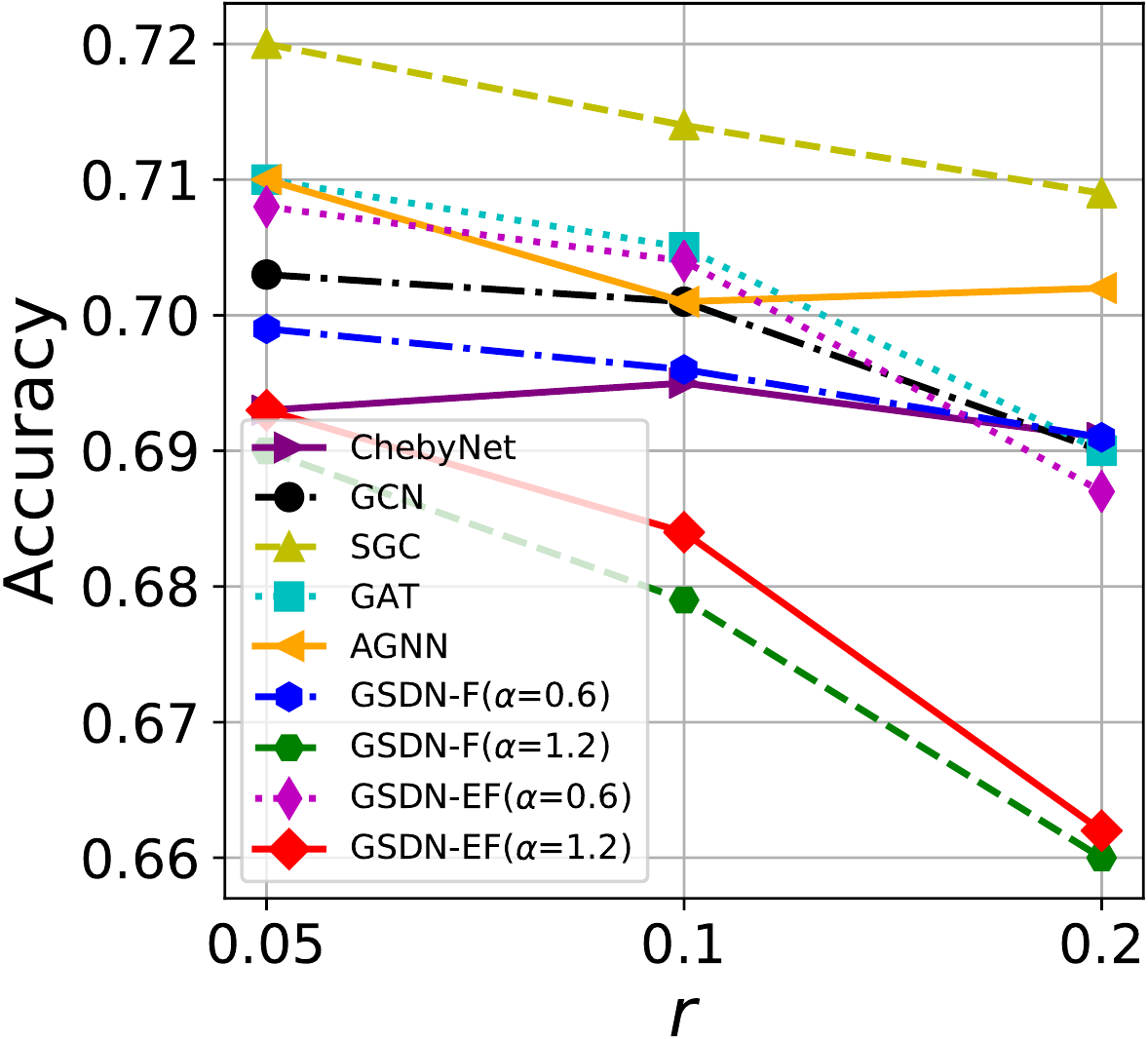}
	}
	\subfigure[CiteSeer w/ node feature noise and edge noise\label{figure3-6}]{
		\includegraphics[width=1.62in]{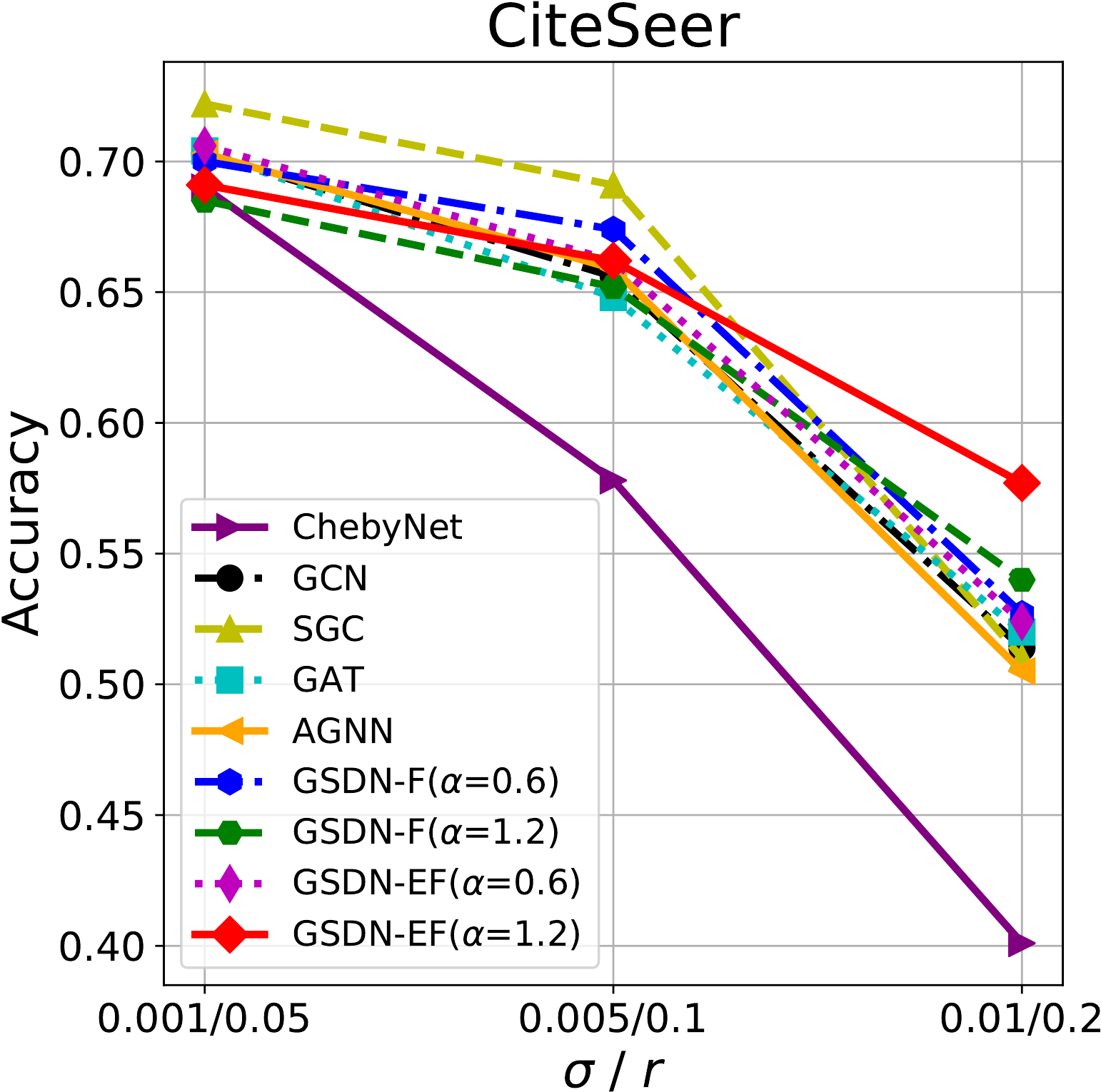}
	}
	\caption{Results of semi-supervised node classification}
	\label{figure3}
\end{figure}

\paragraph{Results on graphs with noise.} We used Cora and CiteSeer by adding noise to node features and/or edges. The results on more datasets (i.e., Pubmed and Coauthor CS) are given in Appendix~\ref{app:noise}. We first normalized the node features and then added Gaussian noise with $\mu=0$ and $\sigma=\{0.001, 0.005, 0.01\}$ into the node features, and report the results in Figures~\ref{figure3-1} and~\ref{figure3-4}. We randomly added or removed edges with noise ratio $r=\frac{\#\text{noisy edges}}{\#\text{original edges}}=\{0.05, 0.1, 0.2\}$, and report the results in Figures~\ref{figure3-2} and~\ref{figure3-5}. We also added Gaussian noise into node features with $\mu=0$ and $\alpha=\{0.001, 0.005, 0.01\}$ and randomly added or removed edges with $r=\{0.05, 0.1, 0.2\}$, and report the results in Figures~\ref{figure3-3} and~\ref{figure3-6}.

Figures~\ref{figure3-1} and~\ref{figure3-4} show that the performance of GCN, SGC and GSDN-F ($\alpha=0.6$) is close to each other and they significantly outperform ChebyNet in the noisy node feature cases. On the one hand, the results indicate that the performance of GCN, SGC and GSDN-F ($\alpha=0.6$) on node classification benefits from their superior ability on node feature denoising and smoothing, while ChebyNet performs the worst as it does not work well for both node feature denoising and smoothing. On the other hand, the results in Section~\ref{exp:denoising-smoothing} show that GSDN-F ($\alpha=0.6$) is better on node feature denoising, while GCN and SGC are better on node feature smoothing. As a result, GSDN-F ($\alpha=0.6$) outperforms GCN and SGC on node classification for graphs with noisy node features.

Figures~\ref{figure3-2} and~\ref{figure3-5} show that the performance of GAT, AGNN and GSDN-EF ($\alpha=0.6$) is superior than that of the other methods. The performance of GSDN-EF ($\alpha=0.6$) is close to that GAT and AGNN, indicating that the linear edge denoising mechanism used in GSDN-EF is sufficient to compete with the graph attentions of GAT and AGNN. The results also show that GSDN-F ($\alpha=1.2$) and GSDN-EF ($\alpha=1.2$) do not work well in the noisy edge cases. The reason is that $\alpha>1$ is suggested for graphs that have node features with much noise, as discussed in Section~\ref{section4-1}, and thus not suitable for the cases here with only noisy edges.

For graphs with both noisy node features and edges, Figures~\ref{figure3-3} and~\ref{figure3-6} show that when there is little noise (e.g., $\sigma \leq 0.005$ and $r \leq 0.1$), GAT, AGNN and GSDN-EF ($\alpha=0.6)$ outperform the other methods on Cora, while GCN, GSDN-F ($\alpha=0.6$) and GSDN-EF ($\alpha=0.6$) are superior on CiteSeer. However, when there is more noise  (e.g., $\sigma=0.01$ and $r=0.2$), GSDN-EF ($\alpha=1.2$) and GSDN-F ($\alpha=1.2$) outperform the other methods on both Cora and CiteSeer. The results show that when there is little noise, the GNNs benefit from their ability to denoise both noisy node features and noisy edges. But when there is more noise, the ability of denoising noisy node features becomes more important and thus GSDN-EF and GSDN-F become more effective than the other methods.

\paragraph{Analysis.}  The performance of GSDN-F and SGCNs (i.e., ChebyNet, GCN, and SGC) benefits from their ability of node feature denoising and smoothing. It requires different tradeoffs between node feature denoising and smoothing to tackle the cases with different noise levels. For the case when there is no noise, GSDN-F and SGCNs are mainly conducting node feature smoothing. When there is little noise, their good performance comes from both node feature denoising and smoothing. However, when there is more noise, denoising starts to have greater contribution to the performance. The results also show the superior performance of GAT, AGNN and GSDN-EF in the noisy edge cases, which demonstrates their effectiveness in denoising edge weights. Their similar performance on denoising edges indicates the linear edge denoising mechanism used in GSDN-EF is comparable with the graph attentions of GAT and AGNN. In addition, we also observe that GSDN-F and GSDN-EF are more effective on graphs with a lot of noise in the node features. Therefore, our results validate our theoretical findings in Section~\ref{sec:analysis} as well as the effectiveness of  our models (i.e., GSDN-F and GSDN-EF) designed based on the theoretical results.

\section{Conclusions}\label{sec:conclusions}

To better understand the mechanisms of SGCNs and AGNNs for node classification, we presented a theoretical framework to explain how and why they work from graph signal denoising (GSD) perspectives. Our framework shows that the results of the graph convolutions of ChebyNet, GCN, and SGC on node features are polynomial approximations to the solution of a GSD problem on graphs with noisy node features. This indicates that spectral graph convolutions work as denoising and smoothing node features. Similarly, GAT and AGNN are implicitly solving a GSD problem on graphs with noisy node features and edge weights. Based on the theoretical results, we designed two new models, GSDN-F and GSDN-EF, which work effectively for graphs with noisy node features and/or noisy edges. We validated our results with experiments on node feature denoising and smoothing and semi-supervised node classification.


\clearpage

\bibliography{reference}

\begin{thebibliography}{14}
\providecommand{\natexlab}[1]{#1}
\providecommand{\url}[1]{\texttt{#1}}
\expandafter\ifx\csname urlstyle\endcsname\relax
  \providecommand{\doi}[1]{doi: #1}\else
  \providecommand{\doi}{doi: \begingroup \urlstyle{rm}\Url}\fi

\bibitem[Berger et~al.(2017)Berger, Hannak, and
  Matz]{DBLP:journals/jstsp/BergerHM17}
Berger, P., Hannak, G., and Matz, G.
\newblock Graph signal recovery via primal-dual algorithms for total variation
  minimization.
\newblock \emph{J. Sel. Topics Signal Processing}, 11\penalty0 (6):\penalty0
  842--855, 2017.

\bibitem[Chen et~al.(2015)Chen, Sandryhaila, Moura, and
  Kovacevic]{DBLP:journals/tsp/ChenSMK15}
Chen, S., Sandryhaila, A., Moura, J. M.~F., and Kovacevic, J.
\newblock Signal recovery on graphs: Variation minimization.
\newblock \emph{{IEEE} Trans. Signal Processing}, 63\penalty0 (17):\penalty0
  4609--4624, 2015.

\bibitem[Defferrard et~al.(2016)Defferrard, Bresson, and
  Vandergheynst]{DBLP:conf/nips/DefferrardBV16}
Defferrard, M., Bresson, X., and Vandergheynst, P.
\newblock Convolutional neural networks on graphs with fast localized spectral
  filtering.
\newblock In \emph{Advances in Neural Information Processing Systems 29: Annual
  Conference on Neural Information Processing Systems, December 5-10, 2016,
  Barcelona, Spain}, pp.\  3837--3845, 2016.

\bibitem[Fey \& Lenssen(2019)Fey and Lenssen]{Fey/Lenssen/2019}
Fey, M. and Lenssen, J.~E.
\newblock Fast graph representation learning with {PyTorch Geometric}.
\newblock In \emph{7th International Conference on Learning Representations,
  New Orleans, LA, USA, May 6-9, 2019, Workshop on Representation Learning on
  Graphs and Manifolds}, 2019.

\bibitem[Hamilton et~al.(2017)Hamilton, Ying, and
  Leskovec]{DBLP:conf/nips/HamiltonYL17}
Hamilton, W.~L., Ying, Z., and Leskovec, J.
\newblock Inductive representation learning on large graphs.
\newblock In \emph{Advances in Neural Information Processing Systems 30: Annual
  Conference on Neural Information Processing Systems, 4-9 December 2017, Long
  Beach, CA, {USA}}, pp.\  1024--1034, 2017.

\bibitem[Hammond et~al.(2011)Hammond, Vandergheynst, and
  Gribonval]{hammond2011wavelets}
Hammond, D.~K., Vandergheynst, P., and Gribonval, R.
\newblock Wavelets on graphs via spectral graph theory.
\newblock \emph{Applied and Computational Harmonic Analysis}, 30\penalty0
  (2):\penalty0 129--150, 2011.

\bibitem[Henaff et~al.(2015)Henaff, Bruna, and
  LeCun]{DBLP:journals/corr/HenaffBL15}
Henaff, M., Bruna, J., and LeCun, Y.
\newblock Deep convolutional networks on graph-structured data.
\newblock \emph{CoRR}, abs/1506.05163, 2015.

\bibitem[Kipf \& Welling(2017)Kipf and Welling]{DBLP:conf/iclr/KipfW17}
Kipf, T.~N. and Welling, M.
\newblock Semi-supervised classification with graph convolutional networks.
\newblock In \emph{5th International Conference on Learning Representations,
  Toulon, France, April 24-26, 2017, Conference Track Proceedings}, 2017.

\bibitem[Sandryhaila \& Moura(2013)Sandryhaila and
  Moura]{DBLP:journals/tsp/SandryhailaM13}
Sandryhaila, A. and Moura, J. M.~F.
\newblock Discrete signal processing on graphs.
\newblock \emph{{IEEE} Trans. Signal Processing}, 61\penalty0 (7):\penalty0
  1644--1656, 2013.

\bibitem[Sen et~al.(2008)Sen, Namata, Bilgic, Getoor, Gallagher, and
  Eliassi{-}Rad]{DBLP:journals/aim/SenNBGGE08}
Sen, P., Namata, G., Bilgic, M., Getoor, L., Gallagher, B., and Eliassi{-}Rad,
  T.
\newblock Collective classification in network data.
\newblock \emph{{AI} Magazine}, 29\penalty0 (3):\penalty0 93--106, 2008.

\bibitem[Thekumparampil et~al.(2018)Thekumparampil, Wang, Oh, and
  Li]{DBLP:journals/corr/abs-1803-03735}
Thekumparampil, K.~K., Wang, C., Oh, S., and Li, L.
\newblock Attention-based graph neural network for semi-supervised learning.
\newblock \emph{CoRR}, abs/1803.03735, 2018.

\bibitem[Velickovic et~al.(2018)Velickovic, Cucurull, Casanova, Romero,
  Li{\`{o}}, and Bengio]{DBLP:conf/iclr/VelickovicCCRLB18}
Velickovic, P., Cucurull, G., Casanova, A., Romero, A., Li{\`{o}}, P., and
  Bengio, Y.
\newblock Graph attention networks.
\newblock In \emph{6th International Conference on Learning Representations,
  Vancouver, BC, Canada, April 30 - May 3, 2018, Conference Track Proceedings},
  2018.

\bibitem[Wu et~al.(2019)Wu, Jr., Zhang, Fifty, Yu, and
  Weinberger]{DBLP:conf/icml/WuSZFYW19}
Wu, F., Jr., A. H.~S., Zhang, T., Fifty, C., Yu, T., and Weinberger, K.~Q.
\newblock Simplifying graph convolutional networks.
\newblock In \emph{Proceedings of the 36th International Conference on Machine
  Learning, 9-15 June 2019, Long Beach, California, {USA}}, pp.\  6861--6871,
  2019.

\bibitem[Zitnik \& Leskovec(2017)Zitnik and
  Leskovec]{DBLP:journals/bioinformatics/ZitnikL17}
Zitnik, M. and Leskovec, J.
\newblock Predicting multicellular function through multi-layer tissue
  networks.
\newblock \emph{Bioinformatics}, 33\penalty0 (14):\penalty0 190--198, 2017.

\end{thebibliography}


\begin{thebibliography}{}

\bibitem[Bahdanau et~al., 2015]{DBLP:journals/corr/BahdanauCB15}
Bahdanau, D., Cho, K., and Bengio, Y. (2015).
\newblock Neural machine translation by jointly learning to align and
  translate.
\newblock In {\em 3rd International Conference on Learning Representations, San
  Diego, CA, USA, May 7-9, 2015, Conference Track Proceedings}.

\bibitem[Battaglia et~al., 2018]{DBLP:journals/corr/abs-1806-01261}
Battaglia, P.~W., Hamrick, J.~B., Bapst, V., Sanchez{-}Gonzalez, A., Zambaldi,
  V.~F., Malinowski, M., Tacchetti, A., Raposo, D., Santoro, A., Faulkner, R.,
  G{\"{u}}l{\c{c}}ehre, {\c{C}}., Song, H.~F., Ballard, A.~J., Gilmer, J.,
  Dahl, G.~E., Vaswani, A., Allen, K.~R., Nash, C., Langston, V., Dyer, C.,
  Heess, N., Wierstra, D., Kohli, P., Botvinick, M., Vinyals, O., Li, Y., and
  Pascanu, R. (2018).
\newblock Relational inductive biases, deep learning, and graph networks.
\newblock {\em CoRR}, abs/1806.01261.

\bibitem[Berger et~al., 2017]{DBLP:journals/jstsp/BergerHM17}
Berger, P., Hannak, G., and Matz, G. (2017).
\newblock Graph signal recovery via primal-dual algorithms for total variation
  minimization.
\newblock {\em J. Sel. Topics Signal Processing}, 11(6):842--855.

\bibitem[Chen et~al., 2015]{DBLP:journals/tsp/ChenSMK15}
Chen, S., Sandryhaila, A., Moura, J. M.~F., and Kovacevic, J. (2015).
\newblock Signal recovery on graphs: Variation minimization.
\newblock {\em {IEEE} Trans. Signal Processing}, 63(17):4609--4624.

\bibitem[Defferrard et~al., 2016]{DBLP:conf/nips/DefferrardBV16}
Defferrard, M., Bresson, X., and Vandergheynst, P. (2016).
\newblock Convolutional neural networks on graphs with fast localized spectral
  filtering.
\newblock In {\em Advances in Neural Information Processing Systems 29: Annual
  Conference on Neural Information Processing Systems, December 5-10, 2016,
  Barcelona, Spain}, pages 3837--3845.

\bibitem[Fey and Lenssen, 2019]{Fey/Lenssen/2019}
Fey, M. and Lenssen, J.~E. (2019).
\newblock Fast graph representation learning with {PyTorch Geometric}.
\newblock In {\em 7th International Conference on Learning Representations, New
  Orleans, LA, USA, May 6-9, 2019, Workshop on Representation Learning on
  Graphs and Manifolds}.

\bibitem[Gordon and Tibshirani, 2012]{gordon2012karush}
Gordon, G. and Tibshirani, R. (2012).
\newblock Karush-kuhn-tucker conditions.
\newblock {\em Optimization}, 10(725/36):725.

\bibitem[Hamilton et~al., 2017]{DBLP:conf/nips/HamiltonYL17}
Hamilton, W.~L., Ying, Z., and Leskovec, J. (2017).
\newblock Inductive representation learning on large graphs.
\newblock In {\em Advances in Neural Information Processing Systems 30: Annual
  Conference on Neural Information Processing Systems, 4-9 December 2017, Long
  Beach, CA, {USA}}, pages 1024--1034.

\bibitem[Hammond et~al., 2011]{hammond2011wavelets}
Hammond, D.~K., Vandergheynst, P., and Gribonval, R. (2011).
\newblock Wavelets on graphs via spectral graph theory.
\newblock {\em Applied and Computational Harmonic Analysis}, 30(2):129--150.

\bibitem[Henaff et~al., 2015]{DBLP:journals/corr/HenaffBL15}
Henaff, M., Bruna, J., and LeCun, Y. (2015).
\newblock Deep convolutional networks on graph-structured data.
\newblock {\em CoRR}, abs/1506.05163.

\bibitem[Kipf and Welling, 2017]{DBLP:conf/iclr/KipfW17}
Kipf, T.~N. and Welling, M. (2017).
\newblock Semi-supervised classification with graph convolutional networks.
\newblock In {\em 5th International Conference on Learning Representations,
  Toulon, France, April 24-26, 2017, Conference Track Proceedings}.

\bibitem[LeCun and Bengio, 1995]{lecun1995convolutional}
LeCun, Y. and Bengio, Y. (1995).
\newblock Convolutional networks for images, speech, and time series.
\newblock {\em The handbook of brain theory and neural networks},
  3361(10):1995.

\bibitem[LeCun et~al., 2015]{DBLP:journals/nature/LeCunBH15}
LeCun, Y., Bengio, Y., and Hinton, G.~E. (2015).
\newblock Deep learning.
\newblock {\em Nature}, 521(7553):436--444.

\bibitem[Li et~al., 2018]{DBLP:conf/aaai/LiHW18}
Li, Q., Han, Z., and Wu, X. (2018).
\newblock Deeper insights into graph convolutional networks for semi-supervised
  learning.
\newblock In {\em Proceedings of the 32nd Conference on Artificial
  Intelligence, New Orleans, Louisiana, USA, February 2-7, 2018}, pages
  3538--3545.

\bibitem[NT and Maehara, 2019]{DBLP:journals/corr/abs-1905-09550}
NT, H. and Maehara, T. (2019).
\newblock Revisiting graph neural networks: All we have is low-pass filters.
\newblock {\em CoRR}, abs/1905.09550.

\bibitem[Sandryhaila and Moura, 2013]{DBLP:journals/tsp/SandryhailaM13}
Sandryhaila, A. and Moura, J. M.~F. (2013).
\newblock Discrete signal processing on graphs.
\newblock {\em {IEEE} Trans. Signal Processing}, 61(7):1644--1656.

\bibitem[Sandryhaila and Moura, 2014]{DBLP:journals/tsp/SandryhailaM14}
Sandryhaila, A. and Moura, J. M.~F. (2014).
\newblock Discrete signal processing on graphs: Frequency analysis.
\newblock {\em {IEEE} Trans. Signal Processing}, 62(12):3042--3054.

\bibitem[Sen et~al., 2008]{DBLP:journals/aim/SenNBGGE08}
Sen, P., Namata, G., Bilgic, M., Getoor, L., Gallagher, B., and Eliassi{-}Rad,
  T. (2008).
\newblock Collective classification in network data.
\newblock {\em {AI} Magazine}, 29(3):93--106.

\bibitem[Thekumparampil et~al., 2018]{DBLP:journals/corr/abs-1803-03735}
Thekumparampil, K.~K., Wang, C., Oh, S., and Li, L. (2018).
\newblock Attention-based graph neural network for semi-supervised learning.
\newblock {\em CoRR}, abs/1803.03735.

\bibitem[Vaswani et~al., 2017]{DBLP:conf/nips/VaswaniSPUJGKP17}
Vaswani, A., Shazeer, N., Parmar, N., Uszkoreit, J., Jones, L., Gomez, A.~N.,
  Kaiser, L., and Polosukhin, I. (2017).
\newblock Attention is all you need.
\newblock In {\em Advances in Neural Information Processing Systems 30: Annual
  Conference on Neural Information Processing Systems, 4-9 December 2017, Long
  Beach, CA, {USA}}, pages 5998--6008.

\bibitem[Velickovic et~al., 2018]{DBLP:conf/iclr/VelickovicCCRLB18}
Velickovic, P., Cucurull, G., Casanova, A., Romero, A., Li{\`{o}}, P., and
  Bengio, Y. (2018).
\newblock Graph attention networks.
\newblock In {\em 6th International Conference on Learning Representations,
  Vancouver, BC, Canada, April 30 - May 3, 2018, Conference Track Proceedings}.

\bibitem[Wu et~al., 2019a]{DBLP:conf/icml/WuSZFYW19}
Wu, F., Jr., A. H.~S., Zhang, T., Fifty, C., Yu, T., and Weinberger, K.~Q.
  (2019a).
\newblock Simplifying graph convolutional networks.
\newblock In {\em Proceedings of the 36th International Conference on Machine
  Learning, 9-15 June 2019, Long Beach, California, {USA}}, pages 6861--6871.

\bibitem[Wu et~al., 2019b]{DBLP:journals/corr/abs-1901-00596}
Wu, Z., Pan, S., Chen, F., Long, G., Zhang, C., and Yu, P.~S. (2019b).
\newblock A comprehensive survey on graph neural networks.
\newblock {\em CoRR}, abs/1901.00596.

\bibitem[Xu et~al., 2019]{DBLP:conf/iclr/XuHLJ19}
Xu, K., Hu, W., Leskovec, J., and Jegelka, S. (2019).
\newblock How powerful are graph neural networks?
\newblock In {\em 7th International Conference on Learning Representations, New
  Orleans, LA, USA, May 6-9, 2019, Conference Track Proceedings}.

\bibitem[Ying et~al., 2019]{DBLP:conf/nips/YingBYZL19}
Ying, Z., Bourgeois, D., You, J., Zitnik, M., and Leskovec, J. (2019).
\newblock Gnnexplainer: Generating explanations for graph neural networks.
\newblock In {\em Advances in Neural Information Processing Systems 32: Annual
  Conference on Neural Information Processing Systems, 8-14 December 2019,
  Vancouver, BC, Canada}, pages 9240--9251.

\bibitem[Zitnik and Leskovec, 2017]{DBLP:journals/bioinformatics/ZitnikL17}
Zitnik, M. and Leskovec, J. (2017).
\newblock Predicting multicellular function through multi-layer tissue
  networks.
\newblock {\em Bioinformatics}, 33(14):190--198.

\end{thebibliography}
\bibliographystyle{apalike}

\clearpage

\appendix

	\section{Theorem} \label{app:theorem}
	\subsection{Proof of Proposition 1}\label{app:prop1}
	\begin{proof}
		For ChebyNet, by Eq.\ref{e13} we have
		\begin{align}
		\sum_{k=0}^K \theta_kT_k(\tilde{\mathbf{L}}_n) = {} & \theta_0T_0(\tilde{\mathbf{L}}_n) + \theta_1T_1(\tilde{\mathbf{L}}_n) + \theta_2T_2(\tilde{\mathbf{L}}_n) + \dots + \theta_KT_K(\tilde{\mathbf{L}}_n) \notag \\
		= {} & \theta_0\mathbf{I}_N + \theta_1\tilde{\mathbf{L}}_n + \theta_2(2\tilde{\mathbf{L}}_n^2 - \mathbf{I}_N) + \dots + \theta_K\left(2\tilde{\mathbf{L}}_nT_{K-1}(\tilde{\mathbf{L}}_n) - T_{K-2}(\tilde{\mathbf{L}}_n)\right) \notag \\
		\stackrel{(a)}{=} {} & \theta_0^{'}\mathbf{I}_N + \theta_1^{'}\tilde{\mathbf{L}}_n + \theta_2^{'}\tilde{\mathbf{L}}_n^2 + \dots + \theta_K^{'}\tilde{\mathbf{L}}_n^K \notag \\
		= {} & \theta_0^{'}\mathbf{I}_N + \theta_1^{'}(\frac{2}{\lambda_{\text{max}}}\mathbf{L}_n - \mathbf{I}_N) + \theta_2^{'}(\frac{2}{\lambda_{\text{max}}}\mathbf{L}_n - \mathbf{I}_N)^2 + \dots + \theta_K^{'}(\frac{2}{\lambda_{\text{max}}}\mathbf{L}_n - \mathbf{I}_N)^K \notag \\
		\stackrel{(a)}{=} {} & \theta_0^{''}\mathbf{I}_N + \theta_1^{''}\mathbf{A}_n + \theta_2^{''}\mathbf{A}_n^2 + \dots + \theta_K^{''}\mathbf{A}_n^K. \label{e20}
		\end{align}
		\noindent where $\stackrel{(a)}{=}$ indicates that we re-parametrize the coefficients. Eq.\ref{e4} and Eq.\ref{e20} show that the result of ChebyNet graph convolution on node features $\mathbf{X}$ is a $K$-order polynomial approximation of $\hat{\mathbf{X}}^*$.
		
		For SGC, when $K \geq 2$, by Eq.\ref{e15} we have:
		\begin{align}
		\tilde{\mathbf{A}}^K = {} & \left(\tilde{\mathbf{D}}^{-\frac{1}{2}}(\mathbf{I}_N + \mathbf{A})\tilde{\mathbf{D}}^{-\frac{1}{2}}\right)^K \notag \\
		= {} & \left(\tilde{\mathbf{D}}^{-1} + \tilde{\mathbf{D}}^{-\frac{1}{2}}\mathbf{A}\tilde{\mathbf{D}}^{-\frac{1}{2}}\right)^K \notag \\
		\approx {} &\left(\tilde{\mathbf{D}}^{-1} + \mathbf{D}_r\mathbf{A}_n\right)^K \notag \\
		= {} & \tilde{\mathbf{D}}^{-K} + K\tilde{\mathbf{D}}^{-1}\mathbf{D}_r\mathbf{A}_n + \frac{K(K-1)}{2!}(\tilde{\mathbf{D}}^{-1}\mathbf{D}_r)^2\mathbf{A}_n^2 + \frac{K(K-1)(K-2)}{3!}(\tilde{\mathbf{D}}^{-1}\mathbf{D}_r)^3\mathbf{A}_n^3 \notag \\
		& + \dots + (\tilde{\mathbf{D}}^{-1}\mathbf{D}_r)^K\mathbf{A}_n^K, \label{e21}
		\end{align}
		\noindent where $\mathbf{D}_r = \text{diag}(d_1 / \tilde{d}_1, \dots, d_N / \tilde{d}_N)$ and $d_i / \tilde{d}_i \leq 1$. Therefore, by Eq.\ref{e4} and Eq.\ref{e21} the result of SGC on node features is a $K$-order polynomial approximation of $\hat{\mathbf{X}}^*$. Note that GSDN-F and SGC are somewhat similar in the form of approximating the optimal solution of Problem 1. However, the approximation of GSDN-F is better than that of SGC. The reason is that SGC does not well approximate the coefficients of the Taylor expansion of the optimal solution. In contrast, GSDN-F directly uses the Taylor expansion of the optimal solution as its convolution kernel and thus the gap between the results of GSDN-F and the optimal solution should be smaller than the gap between the results of SGC and the optimal solution. SGC can be viewed as a degenerate form of GSDN-F.
		
		For GCN, $K = 1$. Then $\tilde{\mathbf{A}} \approx \tilde{\mathbf{D}}^{-1} + \mathbf{D}_r\mathbf{A}_n$. Therefore, the result of GCN on node features is a first-order polynomial approximation of $\hat{\mathbf{X}}^*$.
	\end{proof}
	\subsection{Proof of Proposition 3}\label{app:prop3}
	\begin{proof}
		Let $\mathbf{H} = (1 - \alpha)\sum_{k=0}^K(\alpha\mathbf{A}_n)^k$, when $0 \leq \alpha < 1$ and $K$ is large, we obtain:
		\begin{align}
		\mathbf{H} = {} & (1 - \alpha)\sum_{k=0}^K(\alpha\mathbf{A}_n)^k \notag \\
		= {} & (1 - \alpha)\sum_{k=0}^K(\alpha\mathbf{Q}\mathbf{\Omega}\mathbf{Q}^\top)^k \notag \\
		= {} & (1 - \alpha)\mathbf{Q}
		\begin{bmatrix}
		\frac{1 - (\alpha\omega_1)^K}{1 - \alpha\omega_1} & & \\
		& \ddots & \\
		& & \frac{1 - (\alpha\omega_N)^K}{1 - \alpha\omega_N}
		\end{bmatrix}
		\mathbf{Q}^\top \notag \\
		\approx {} & (1 - \alpha)\mathbf{Q}
		\begin{bmatrix}
		\frac{1}{1 - \alpha\omega_1} & & \\
		& \ddots & \\
		& & \frac{1}{1 - \alpha\omega_N}
		\end{bmatrix}
		\mathbf{Q}^\top \notag \\
		= {} & \mathbf{Q}
		\begin{bmatrix}
		1 - \frac{1 - \omega_1}{\frac{1}{\alpha} - \omega_1} & & \\
		& \ddots & \\
		& & 1 - \frac{1 - \omega_N}{\frac{1}{\alpha} - \omega_N} \\
		\end{bmatrix}
		\mathbf{Q}^\top. \label{e22}
		\end{align}
		\noindent where $\mathbf{A}_n = \mathbf{Q}\mathbf{\Omega}\mathbf{Q}^\top$ is the eigen-decomposition of $\mathbf{A}_n$, $\mathbf{Q}$ is the matrix of eigenvectors of $\mathbf{A}_n$, $\mathbf{\Omega}$ is its diagonal matrix of eigenvalues and $0 \leq \omega_1 \leq \omega_2 \leq \dots \leq \omega_N \leq 1$ are the eigenvalues. Then we have:
		\begin{align}
		\textbf{Var}(\hat{\mathbf{x}}^* | \alpha) = {} &\mathbb{E}\left[||\hat{\mathbf{x}}^* - \mathbb{E}[\hat{\mathbf{x}}^*]||_2^2\right] \notag \\
		= {} &\mathbb{E}\left[||\mathbf{Hx} - \mathbf{H}\hat{\mathbf{x}}||_2^2\right] \notag \\
		= {} &\mathbb{E}\left[||\mathbf{H}\mathbf{z}||_2^2\right] \notag \\
		= {} &\textbf{Tr}(\mathbf{H}^2\mathbf{\Sigma}) \notag \\
		\approx {} & \textbf{Tr}(\left(\mathbf{Q}
		\begin{bmatrix}
		1 - \frac{1 - \omega_1}{\frac{1}{\alpha} - \omega_1} & & \\
		& \ddots & \\
		& & 1 - \frac{1 - \omega_N}{\frac{1}{\alpha} - \omega_N} \notag \\
		\end{bmatrix}
		\mathbf{Q}^\top\right)^2\mathbf{\Sigma}) \\
		= {} & \textbf{Tr}(\mathbf{Q}
		\begin{bmatrix}
		(1 - \frac{1 - \omega_1}{\frac{1}{\alpha} - \omega_1})^2 & & \\
		& \ddots & \\
		& & (1 - \frac{1 - \omega_N}{\frac{1}{\alpha} - \omega_N})^2 \\
		\end{bmatrix}
		\mathbf{Q}^\top\mathbf{\Sigma}). \label{e23}
		\end{align}
		By $0 \leq \alpha < 1$, we have $0 < \frac{1 - \omega_i}{\frac{1}{\alpha}  - \omega_i}< 1$ for $1 \leq i \leq N$. Then, increasing $\alpha$ will lead to lower variance. Similarly,
		\begin{align}
		\textbf{Bias}(\hat{\mathbf{x}}^* | \hat{\mathbf{x}}, \alpha)^2 = {} & \mathbb{E}\left[||\mathbb{E}[\hat{\mathbf{x}}^*] - \hat{\mathbf{x}}||_2^2\right] \notag \\
		= {} & \mathbb{E}\left[||\mathbf{H}\hat{\mathbf{x}} - \hat{\mathbf{x}}||_2^2\right] \notag \\
		= {} & ||(\mathbf{H} - \mathbf{I}_N)\hat{\mathbf{x}}||_2^2 \notag \\
		\approx {} & \big|\big|\left(\mathbf{Q}
		\begin{bmatrix}
		1 - \frac{1 - \omega_1}{\frac{1}{\alpha} - \omega_1} & & \\
		& \ddots & \\
		& & 1 - \frac{1 - \omega_N}{\frac{1}{\alpha} - \omega_N} \\
		\end{bmatrix}
		\mathbf{Q}^\top - \mathbf{I}_N\right)\hat{\mathbf{x}}\big|\big|_2^2 \notag \\
		= {} & \big|\big|\mathbf{Q}
		\begin{bmatrix}
		1 - \frac{1 - \omega_1}{\frac{1}{\alpha} - \omega_1} - 1 & & \\
		& \ddots & \\
		& & 1 - \frac{1 - \omega_N}{\frac{1}{\alpha} - \omega_N} - 1 \\
		\end{bmatrix}
		\mathbf{Q}^\top\hat{\mathbf{x}}\big|\big|_2^2 \notag \\
		= {} & \big|\big|\mathbf{Q}
		\begin{bmatrix}
		-\frac{1 - \omega_1}{\frac{1}{\alpha} - \omega_1} & & \\
		& \ddots & \\
		& & -\frac{1 - \omega_N}{\frac{1}{\alpha} - \omega_N} \\
		\end{bmatrix}
		\mathbf{Q}^\top\hat{\mathbf{x}}\big|\big|_2^2 \notag \\
		= {} & \big|\big|\mathbf{Q}
		\begin{bmatrix}
		\frac{1 - \omega_1}{\frac{1}{\alpha} - \omega_1} & & \\
		& \ddots & \\
		& & \frac{1 - \omega_N}{\frac{1}{\alpha} - \omega_N} \\
		\end{bmatrix}
		\mathbf{Q}^\top\hat{\mathbf{x}}\big|\big|_2^2. \label{e24}
		\end{align}
		\noindent Therefore, increasing $\alpha$ will lead to higher bias.
	\end{proof}
\section{Additional Experiments}\label{app:add_exp}

	\subsection{Experiments on Other Graphs with Noise}\label{app:noise}
\begin{figure}[!t]
		\centering
		\subfigure[Pubmed w/ node feature noise\label{figure6-1}]{
			\includegraphics[width=1.62in]{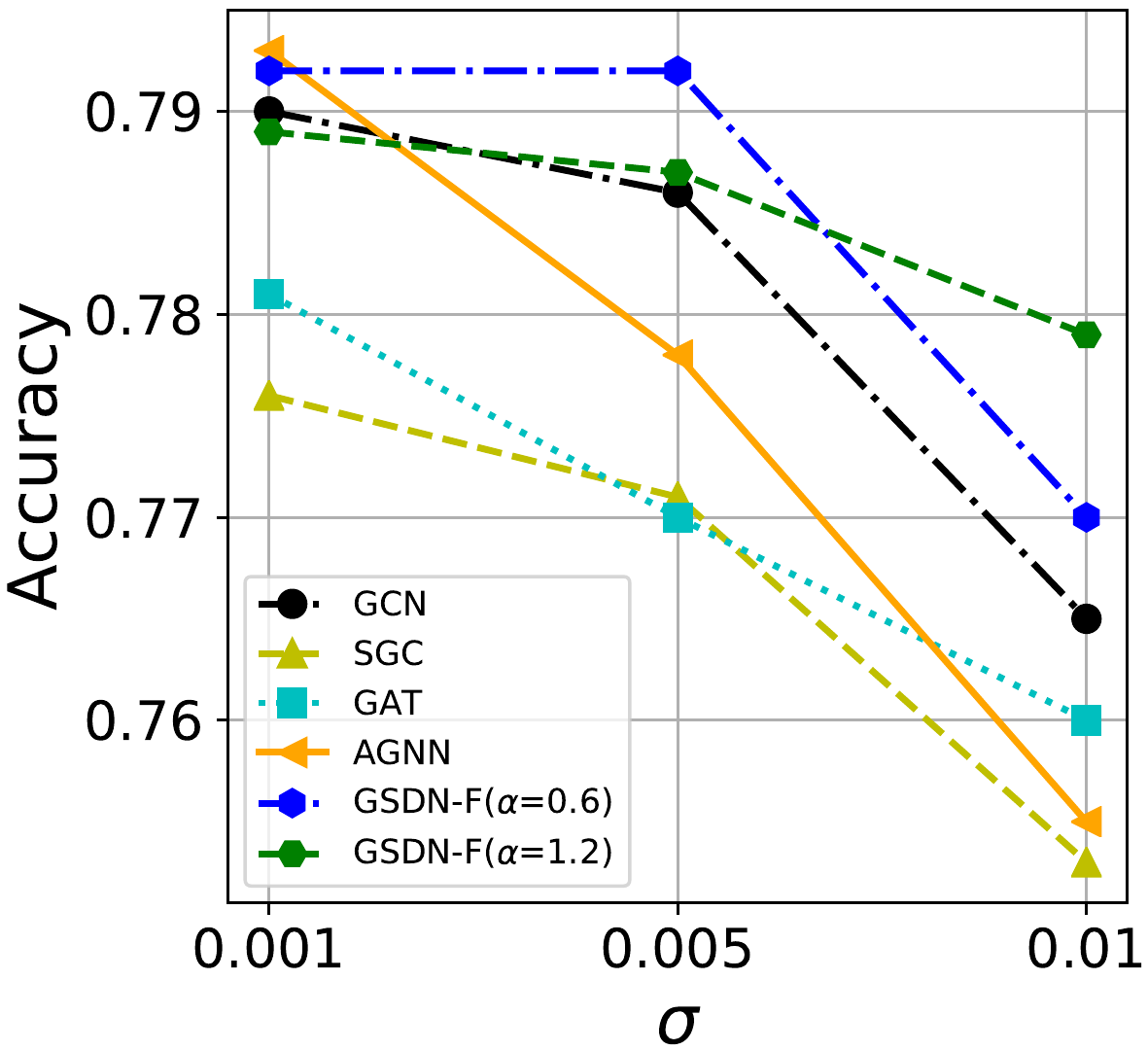}
		}
		\subfigure[Pubmed w/ edge noise\label{figure6-2}]{
			\includegraphics[width=1.62in]{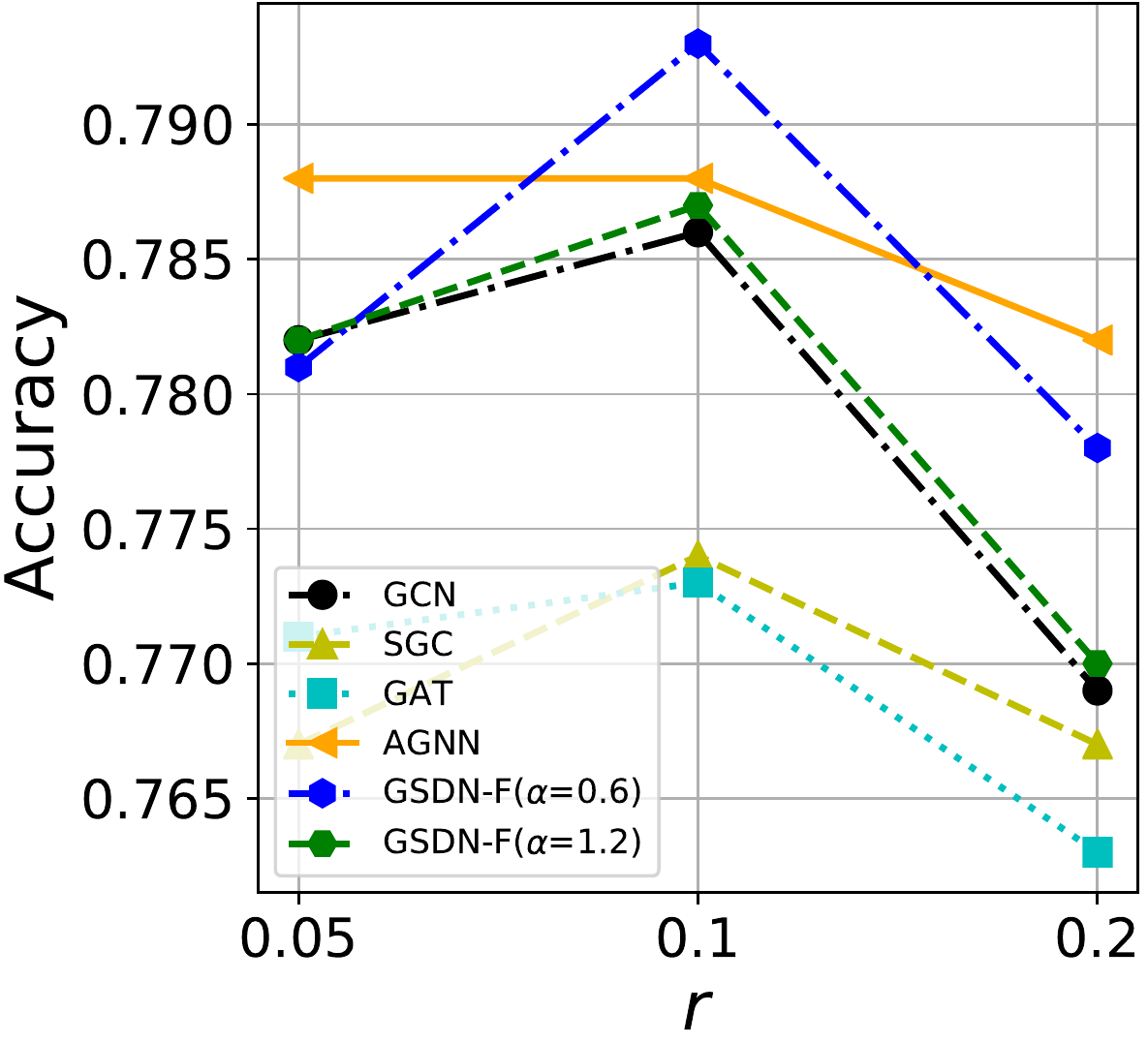}
		}
		\subfigure[Pubmed w/ node feature noise and edge noise\label{figure6-3}]{
			\includegraphics[width=1.63in]{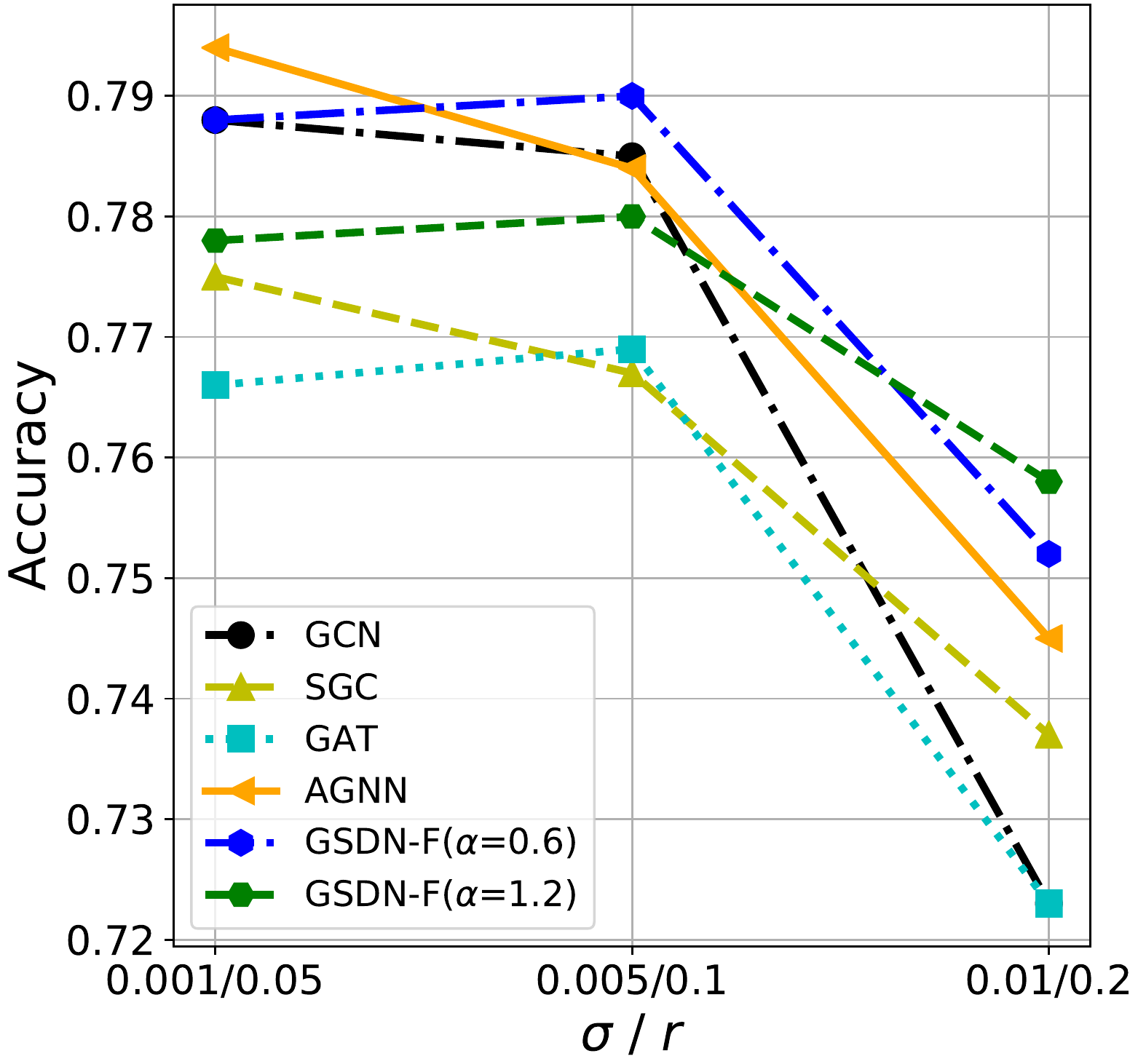}
		}
		\subfigure[Coauthor CS w/ node feature noise\label{figure6-4}]{
			\includegraphics[width=1.62in]{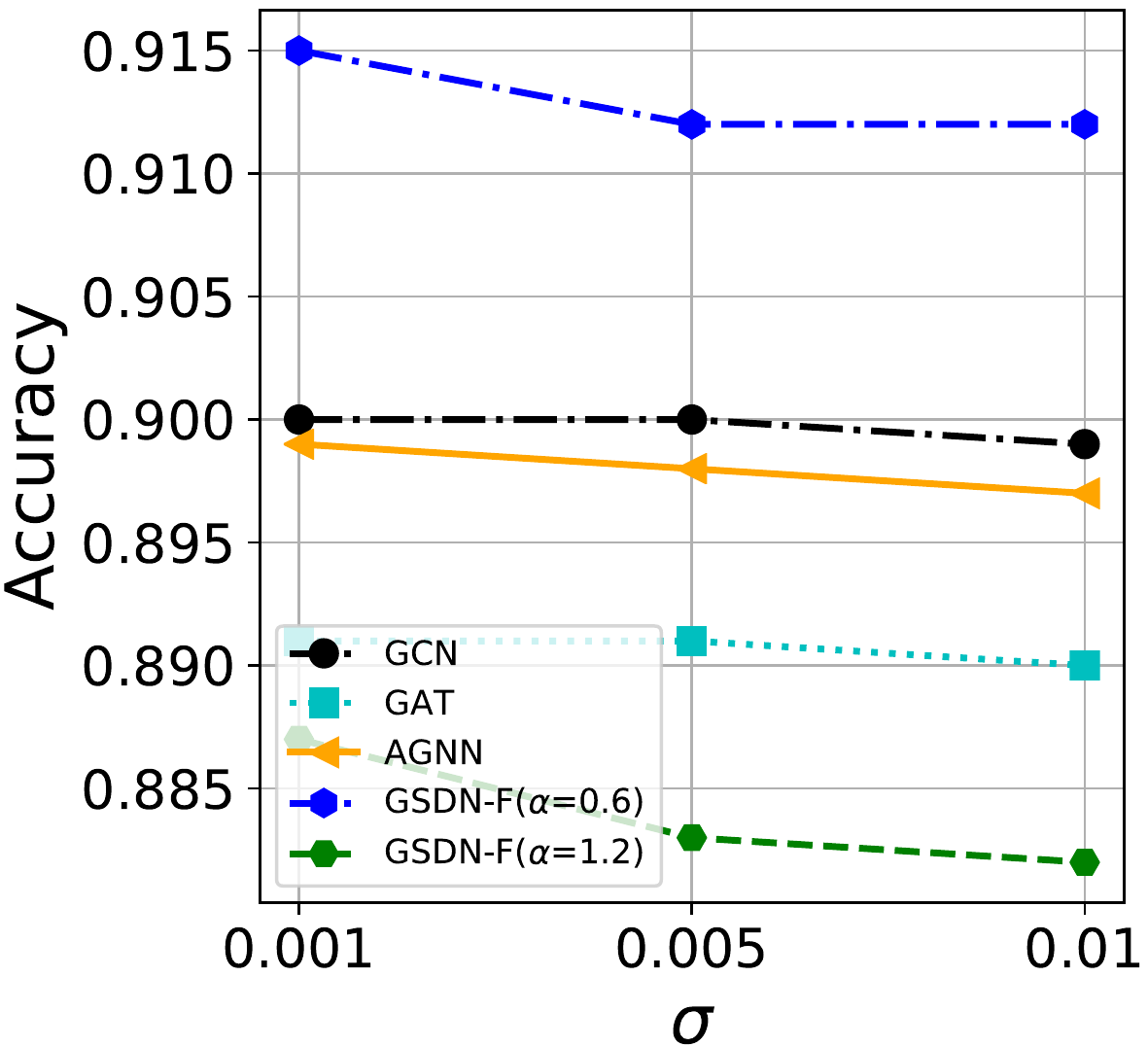}
		}
		\subfigure[Coauthor CS w/ edge noise\label{figure6-5}]{
			\includegraphics[width=1.62in]{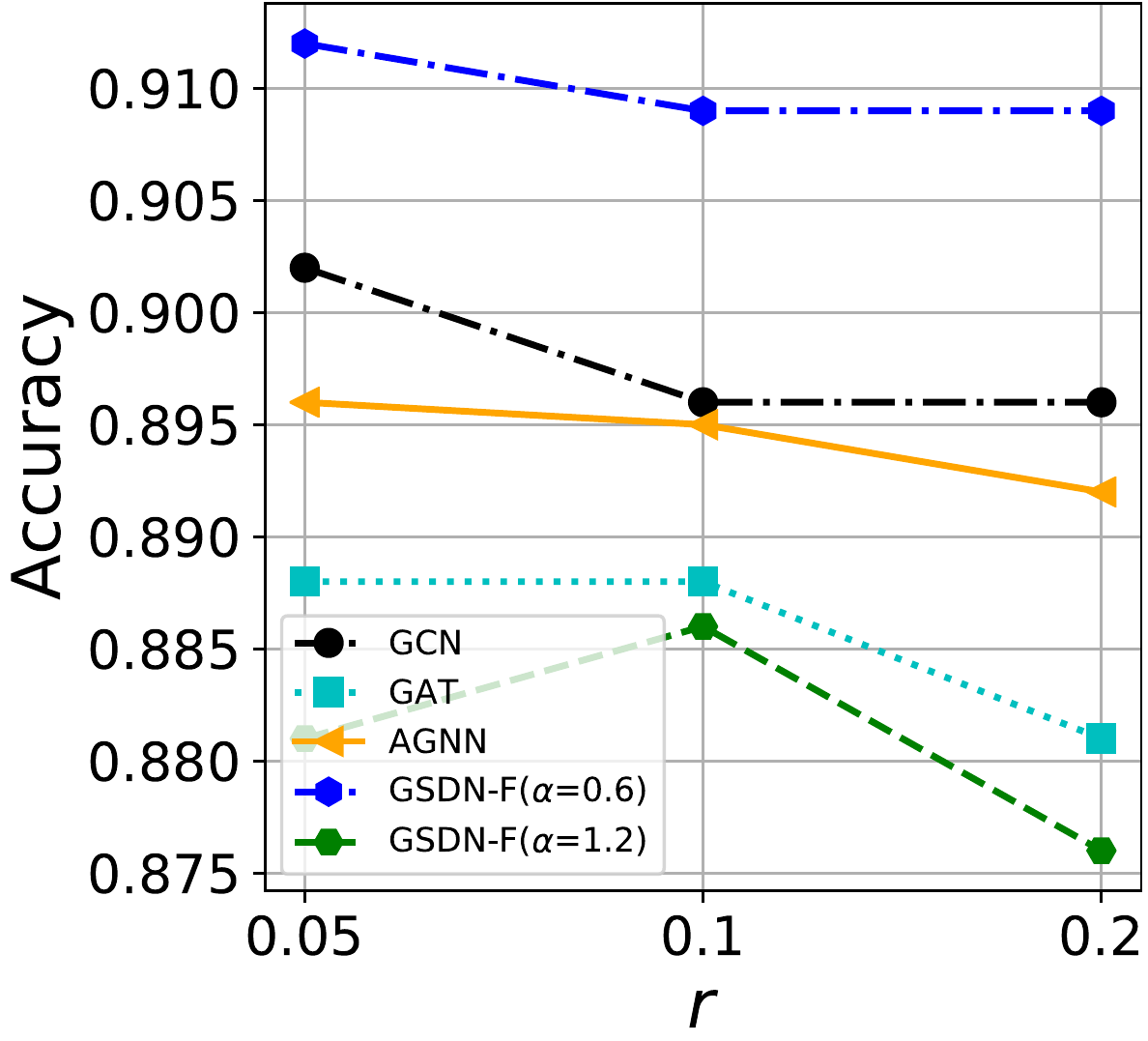}
		}
		\subfigure[Coauthor CS w/ node feature noise and edge noise\label{figure6-6}]{
			\includegraphics[width=1.62in]{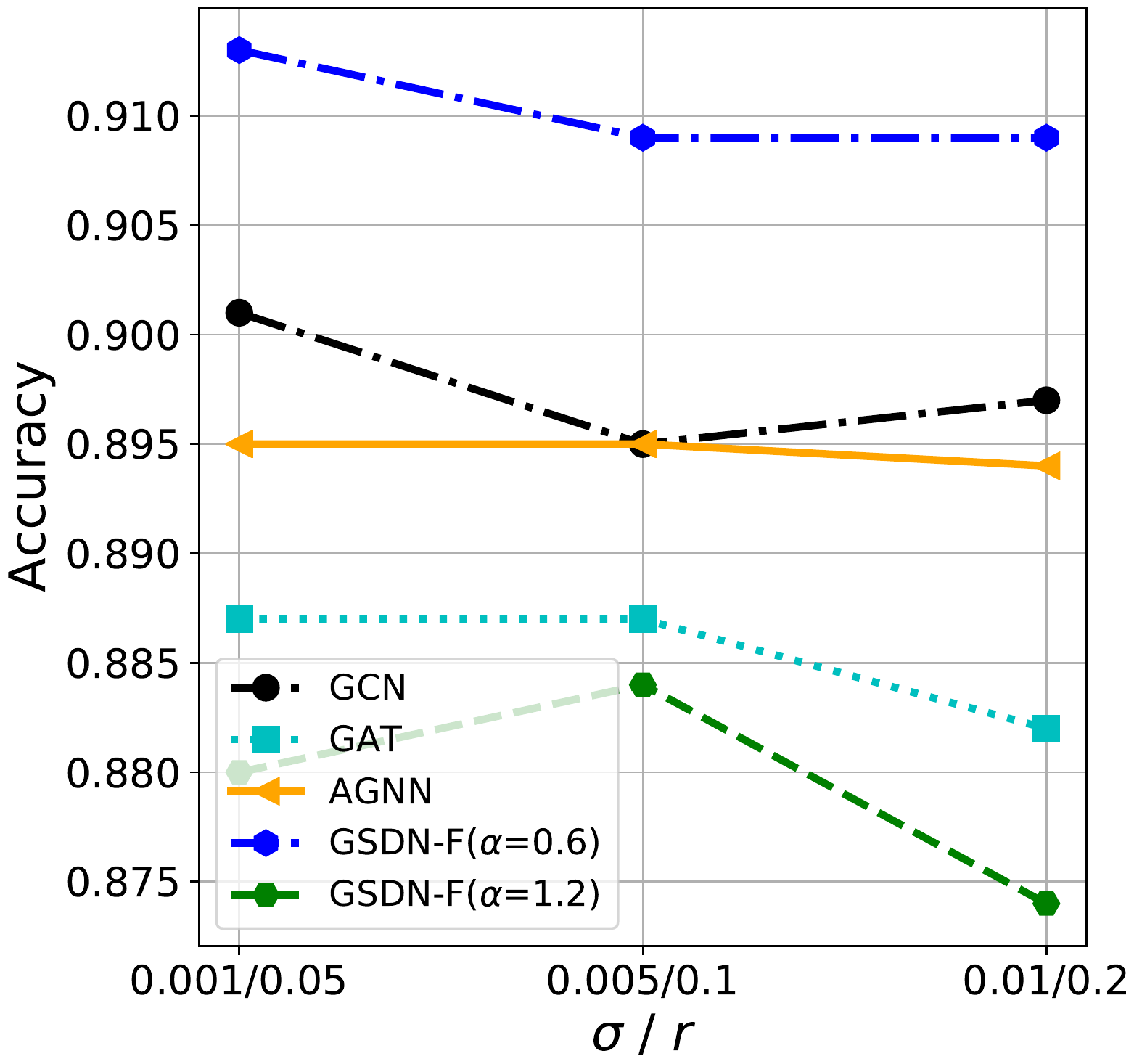}
		}
		\caption{Results of semi-supervised node classification on the Pubmed and Coauthor CS datasets}
		\label{figure6}
	\end{figure}

	We conducted experiments on another two datasets, Pubmed and Coauthor CS, using the same settings on graphs with noise in Section~\ref{exp:classification}. We first normalized the node features and then added Gaussian noise with $\mu=0$ and $\sigma=\{0.001, 0.005, 0.01\}$ into the node features, and report the results in Figures~\ref{figure6-1} and~\ref{figure6-4}. We randomly added or removed edges with noise ratio $r=\frac{\#\text{noisy edges}}{\#\text{original edges}}=\{0.05, 0.1, 0.2\}$, and report the results in Figures~\ref{figure6-2} and~\ref{figure6-5}. We also added Gaussian noise into node features with $\mu=0$ and $\alpha=\{0.001, 0.005, 0.01\}$ and randomly added or removed edges with $r=\{0.05, 0.1, 0.2\}$, and report the results in Figures~\ref{figure6-3} and~\ref{figure6-6}.
	
	Figures~\ref{figure6-1} and~\ref{figure6-4} show that the performance of GSDN-F ($\alpha=0.6$) is better than the other models in the noisy node feature cases. The results on Pubmed and Coauthor CS are similar to those on Cora and CiteSeer reported in Figures~\ref{figure3-1} and~\ref{figure3-4}, which consistently show that GSDN-F ($\alpha=0.6$) outperforms GCN and SGC on node classification for graphs with noisy node features.
	
	Figures~\ref{figure6-2} and~\ref{figure6-5} show that GSDN-F ($\alpha=0.6$) also obtained superior performance in the noisy edge cases on Pubmed and Coauthor CS. The results validate that GSDN-F ($\alpha=0.6$) works well in the noisy edge cases on Pubmed and Coauthor CS.
	
	For graphs with both noisy node features and edges, the results of Figures~\ref{figure6-3} and~\ref{figure6-6} demonstrate that GSDN-F ($\alpha=0.6$) outperforms the models in most cases. Again, GSDN-F ($\alpha=0.6$) has the best performance in the noisy feature and noisy edge cases on Pubmed and Coauthor CS.
	
	Thus, the results in Figure~\ref{figure6} demonstrate the effectiveness of GSDN-F on graphs with noisy features and/or noisy edges, which is similar with the results obtained on Cora and CiteSeer presented in Section~\ref{exp:classification}.

	\subsection{Parameter Sensitivity Tests}\label{app:sensitivity}
	
\begin{figure}[!t]
		\centering
		\subfigure[$\alpha$]{
			\includegraphics[width=2in]{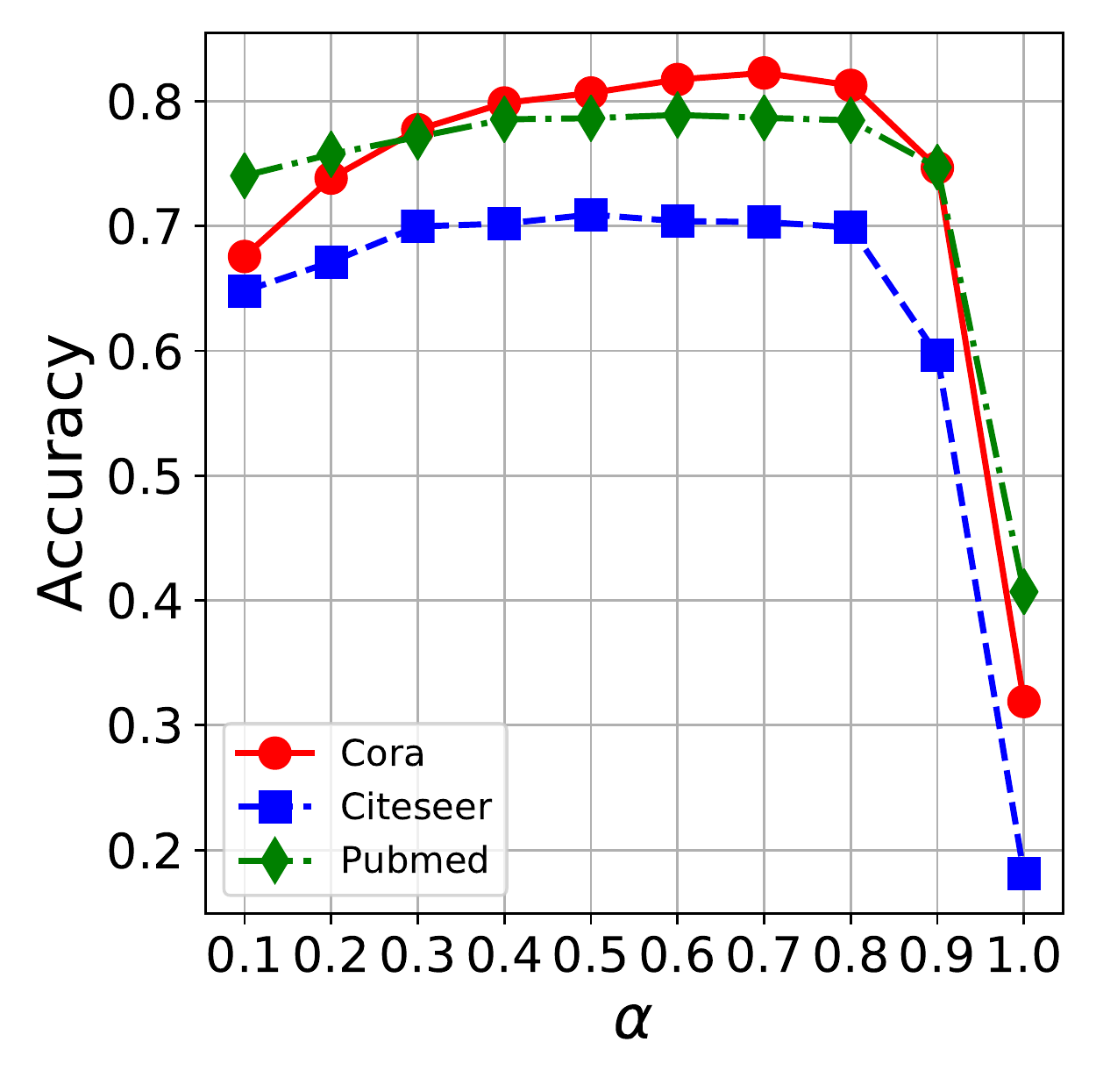}\label{figure7-1}
		}
		\subfigure[$K$]{
			\includegraphics[width=2.01in]{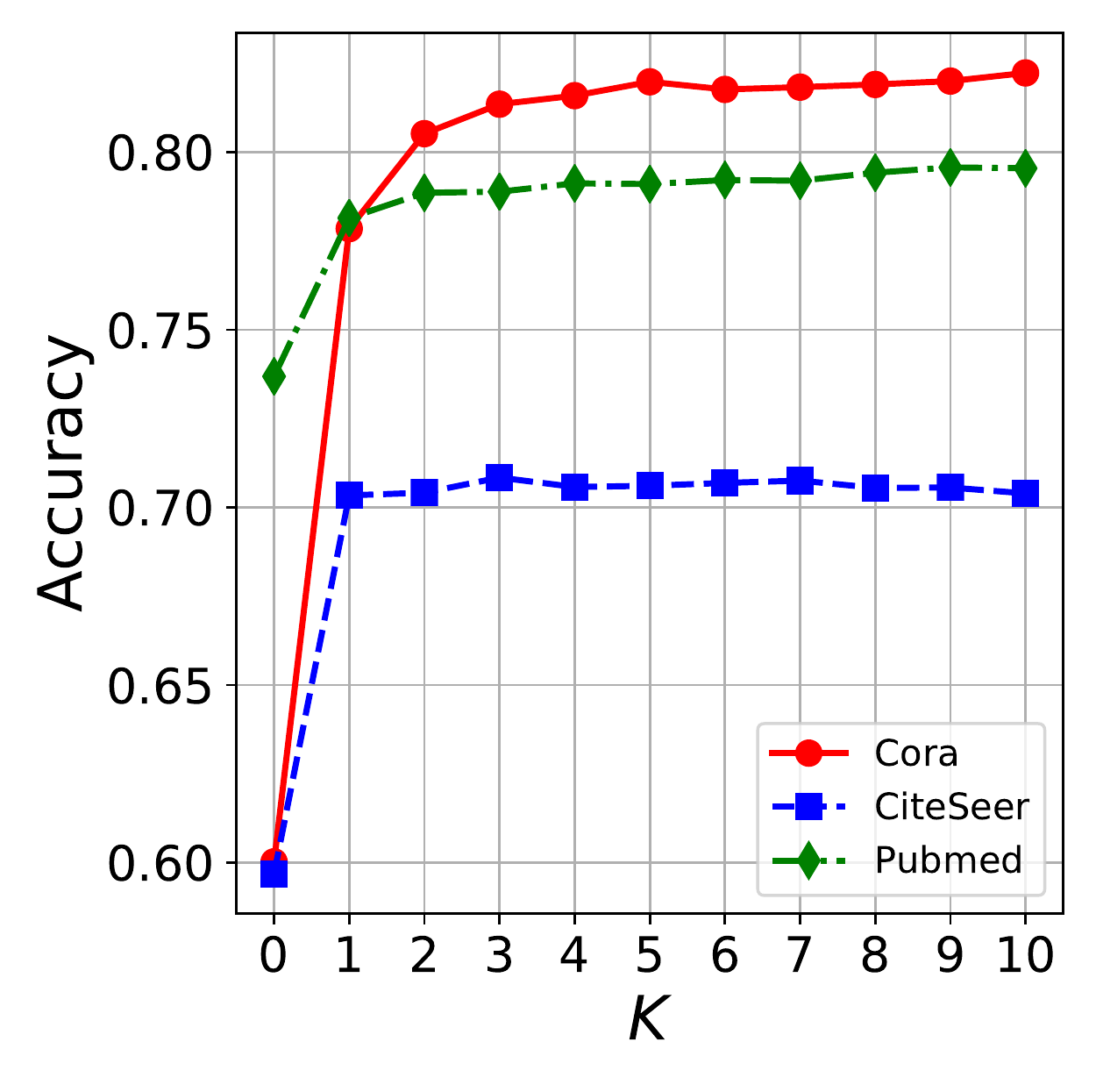}\label{figure7-2}
		}
		\caption{Results of parameter sensitivity}
		\label{figure7}
	\end{figure}

	In this experiment, we studied the sensitivity of $\alpha$ and $K$. The results are reported in Figure~\ref{figure7}. 
	\begin{figure}[!t]
		\centering
		\subfigure[Original graphs]{
			\includegraphics[width=2in]{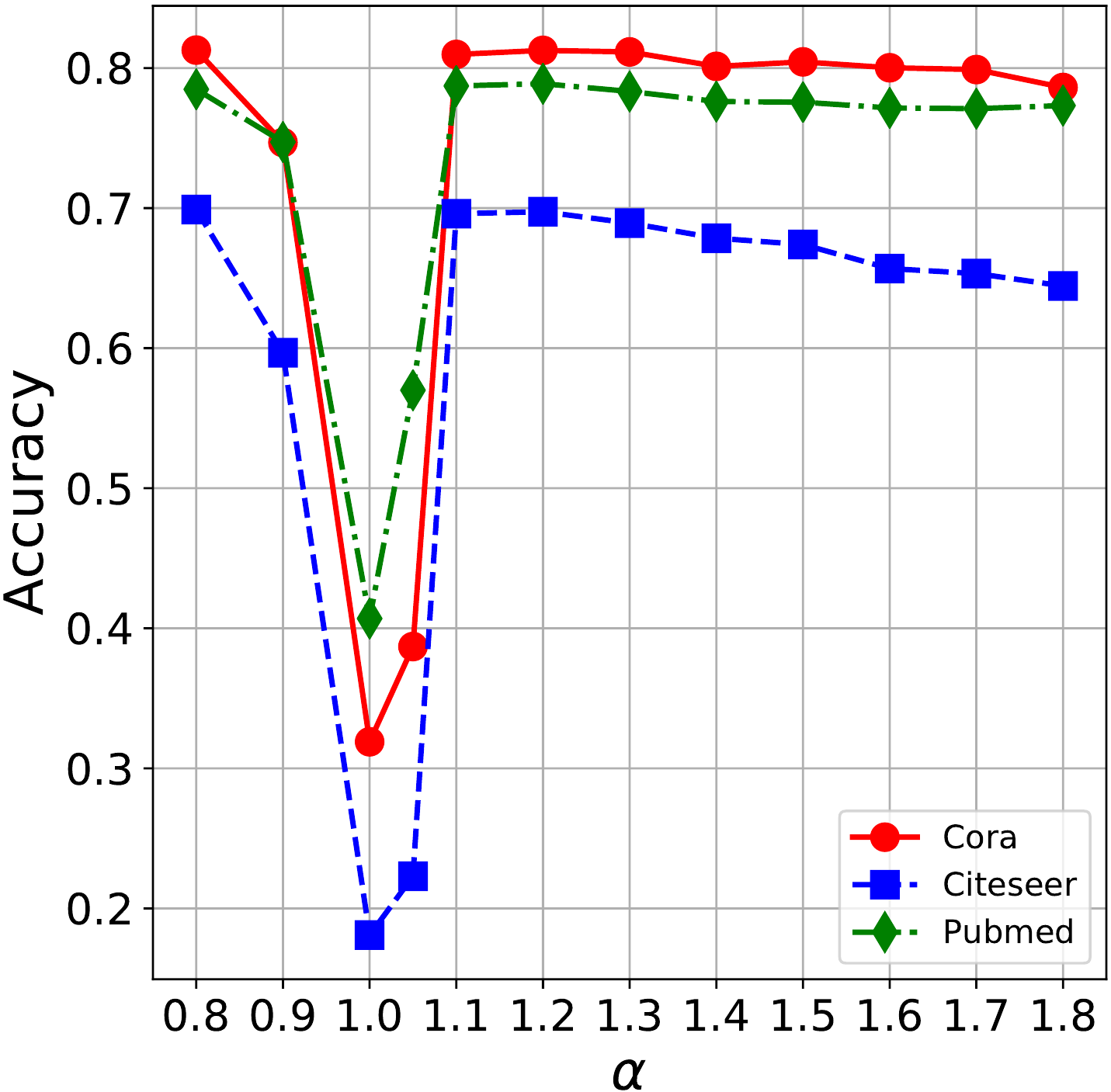}\label{figure8-1}
		}
		\subfigure[Graphs with Gaussian noise ($\mu=0$, $\sigma=0.01$) in node features]{
			\includegraphics[width=2in]{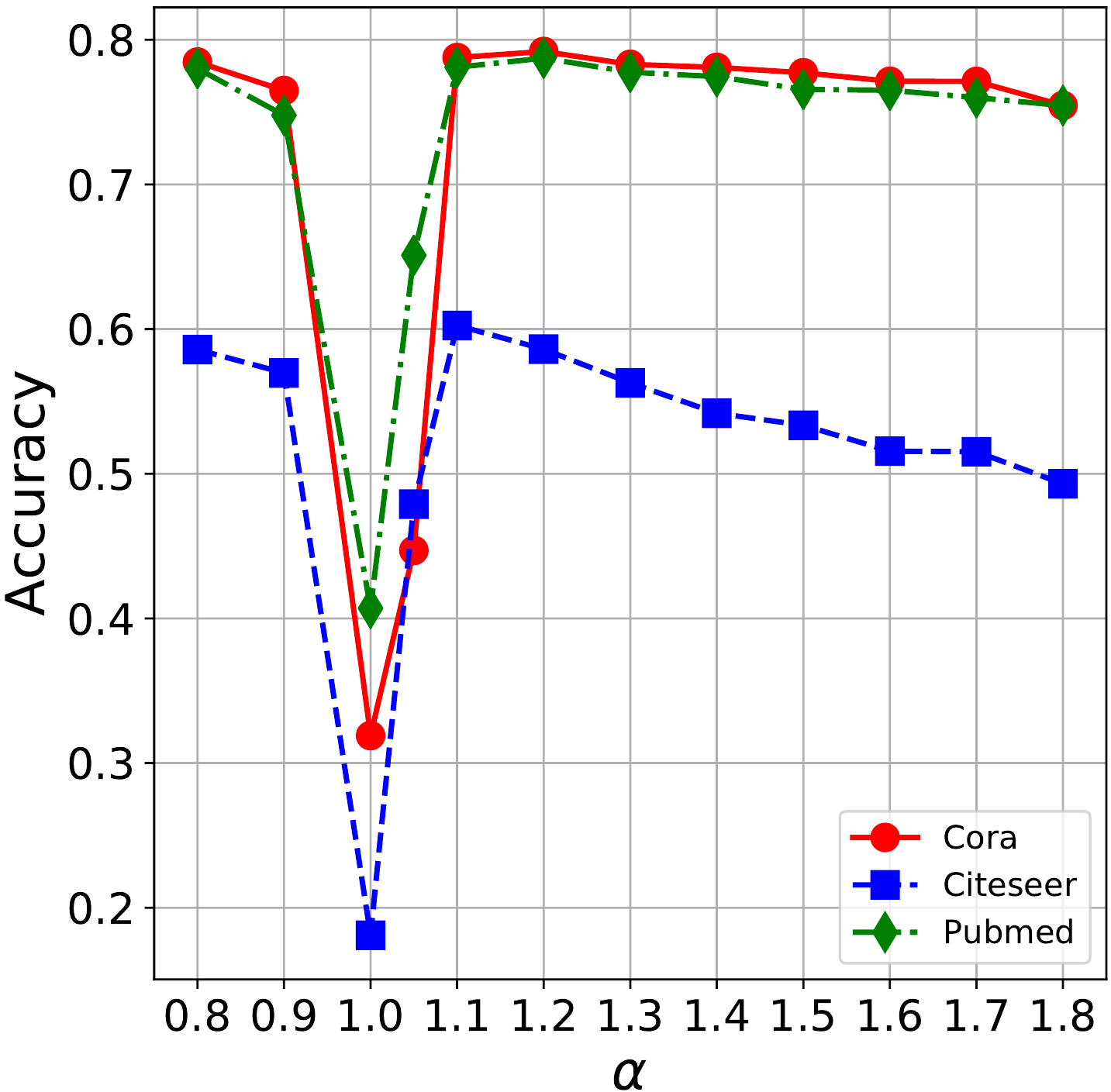}\label{figure8-2}
		}
		\caption{Results of parameter sensitivity of $\alpha$ for $\alpha > 1$}
		\label{figure8}
	\end{figure}

	\paragraph{Variance-bias trade-off by $\alpha$.} When $0 < \alpha \leq 1$, Figure~\ref{figure7-1} shows that as $\alpha$ increases, the performance of GSDN-F improves first and then dramatically degrades when $\alpha$ is larger than 0.8. The result illustrates that the choice of $\alpha$ can seriously affect the performance of node classification  and validates the variance-bias trade-off by $\alpha$ as stated in Proposition~\ref{prop3}.
	
	\paragraph{The effect of polynomial approximation order $K$.} Figure~\ref{figure7-2} shows that the performance of GSDN-F improves as $K$ increases and then becomes stable when $K$ is large enough. This is because the larger the value of $K$ is, the more accurate is the approximation and hence the better performance of GSDN-F is.

	\paragraph{Parameter Sensitivity for $\alpha > 1$.} Here we study the sensitivity of $\alpha$ for $\alpha>1$. We conducted the experiments for semi-supervised node classification on Cora, CiteSeer, and Pubmed with and without Gaussian noise ($\mu=0$, $\sigma=0.01$). The results are reported in Figure~\ref{figure8}.
	
	When $\alpha > 1$, Figure~\ref{figure8} shows that as $\alpha$ increases, the performance of GSDN-F dramatically improves first and slightly degrades when $\alpha$ is larger than 1.2 on both the original graphs and the graphs with Gaussian noise in node features. The results suggest that choosing GSDN-F with $\alpha$ around 1.2 can obtain good performance for graphs with a lot of noise.

\end{document}